\documentclass[pdflatex,sn-apa]{sn-jnl}% APA Reference Style
\pdfoutput=1
\usepackage{algorithm}  
\usepackage{algpseudocode}  
\usepackage{amsmath}  
\usepackage{graphicx}

\theoremstyle{thmstyleone}%
\theoremstyle{thmstyletwo}%

\theoremstyle{thmstylethree}%

\begin{document}

\title[ ]{Towards Adaptive Unknown Authentication for Universal Domain Adaptation by Classifier Paradox}


\author[1]{\fnm{Yunyun} \sur{Wang}}\email{wangyunyun@njupt.edu.cn}

\author[1]{\fnm{Yao} \sur{Liu}}\email{1020041116@njupt.edu.cn}

\author*[2]{\fnm{Songcan} \sur{Chen}}\email{s.chen@nuaa.edu.cn}

\affil[1]{\orgdiv{School of Computer Science and Engineering}, \orgname{Nanjing University of Posts and Telecommunications}, \orgaddress{\city{Nanjing}, \postcode{210046}, \country{China}}}

\affil*[2]{\orgdiv{School of Computer Science and Technology}, \orgname{Nanjing University of Aeronautics and Astronautics}, \orgaddress{\city{Nanjing}, \postcode{210023}, \country{China}}}

\abstract{Universal domain adaptation (UniDA) is a general unsupervised domain adaptation setting, which addresses both domain and label shifts in adaptation. Its main challenge lies in how to identify target samples in unshared or unknown classes. Previous methods commonly strive to depict sample ``confidence'' along with a threshold for rejecting unknowns, and align feature distributions of shared classes across domains. However, it is still hard to pre-specify a ``confidence'' criterion and threshold which are adaptive to various real tasks, and a mis-prediction of unknowns further incurs mis-alignment of features in shared classes. In this paper, we propose a new UniDA method with adaptive \emph{U}nknown \emph{A}uthentication by \emph{C}lassifier \emph{P}aradox (UACP), considering that samples with paradoxical predictions are probably unknowns belonging to none of the source classes. In UACP, a composite classifier is jointly designed with two types of predictors. That is, a multi-class (MC) predictor classifies samples to one of the multiple source classes, while a binary one-vs-all (OVA) predictor further verifies the prediction by MC predictor. Samples with verification failure or paradox are identified as unknowns. Further, instead of feature alignment for shared classes, implicit domain alignment is conducted in output space such that samples across domains share the same decision boundary, though with feature discrepancy. Empirical results validate UACP under both open-set and universal UDA settings.}


\maketitle
\section{Introduction}\label{sec1}
% \cite{ben2010theory,tzeng2014deep,long2015learning,long2017deep}
Unsupervised domain adaptation (UDA) (Ben-David et al., \citeyear{ben2010theory}; Tzeng et al., \citeyear{tzeng2014deep}; Long et al., \citeyear{long2015learning})  aims to adopt a fully-labeled source domain to help the learning of unlabeled target domain. Existing UDA methods mainly attempt to generate domain-invariant representations by reducing the distribution discrepancy across domains with some distance metrics, such as Maximum Mean Discrepancy (MMD) (Tzeng et al., \citeyear{tzeng2014deep}), or by adversarial learning (Ganin et al., \citeyear{ganin2016domain}) between the feature generator and domain discriminator. However, they commonly make a strong assumption that the source and target domains share the same label set, which limits their applicability to many real-world applications.

In real tasks, the label sets from source and target domains are usually different. For example, with the emergence of Big Data (Sagiroglu \& Sinanc, \citeyear{sagiroglu2013big}), large-scale labeled datasets like ImageNet-1K (Russakovsky et al., \citeyear{russakovsky2015imagenet}) and Google Open Images (Krasin et al., \citeyear{krasin2017openimages}) are readily accessible as the source domains, while the target domain may only contain a subset of the source classes, leading to a so-termed Partial Domain Adaptation (PDA) (Cao et al., \citeyear{cao2018partial}). On the other hand, in real open learning scenes, target domains usually have unknown classes not covered in the source domain, leading to a setting of so-called Open-Set Domain Adaptation (OSDA) (Panareda Busto \& Gall, \citeyear{panareda2017open}). The learning purpose is to classify the target data in known classes correctly, while reject data in all unknown classes as ``unknown''. Recently, Universal Domain Adaptation (UniDA) (You et al., \citeyear{you2019universal}), a general learning setting without the need of prior knowledge on label sets across domains, has attracted increasing attention. Obviously, UniDA is a more realistic UDA setting since the target ground-truth is actually not available in real tasks.

A main learning challenge posed in such setting is how to identify the target samples in unshared or unknown classes. Previous methods mainly strive to depict the sample ``confidence'' along with a pre-defined threshold to detect target unknowns, then align distributions of shared classes across domains. The commonly adopted confidence criteria include prediction entropy (Saito et al., \citeyear{saito2020universal}), source similarity (Panareda Busto \& Gall, \citeyear{panareda2017open}), classifier discrepancy (Liang et al., \citeyear{liang2021umad}) and minimum inter-class distance (Saito \& Saenko, \citeyear{saito2021ovanet}), etc. Though with great progress, it is still hard to pre-specify a universal ``confidence'' criterion and threshold that are adaptive to various complicated real tasks. Furthermore, a mis-prediction of unknowns further incurs mis-alignment of features in shared classes, probably leading to negative transfer. To this end, we propose a new UniDA method with adaptive unknown authentication by classifier paradox (UACP). Specifically, the prediction paradox from two types of predictors is adopted in UACP to adaptively identify target unknowns, since samples with paradoxical predictions are probably unknowns belonging to none of the source classes.
\begin{figure}[t]
    \centering
    \includegraphics[width=1.0\textwidth]{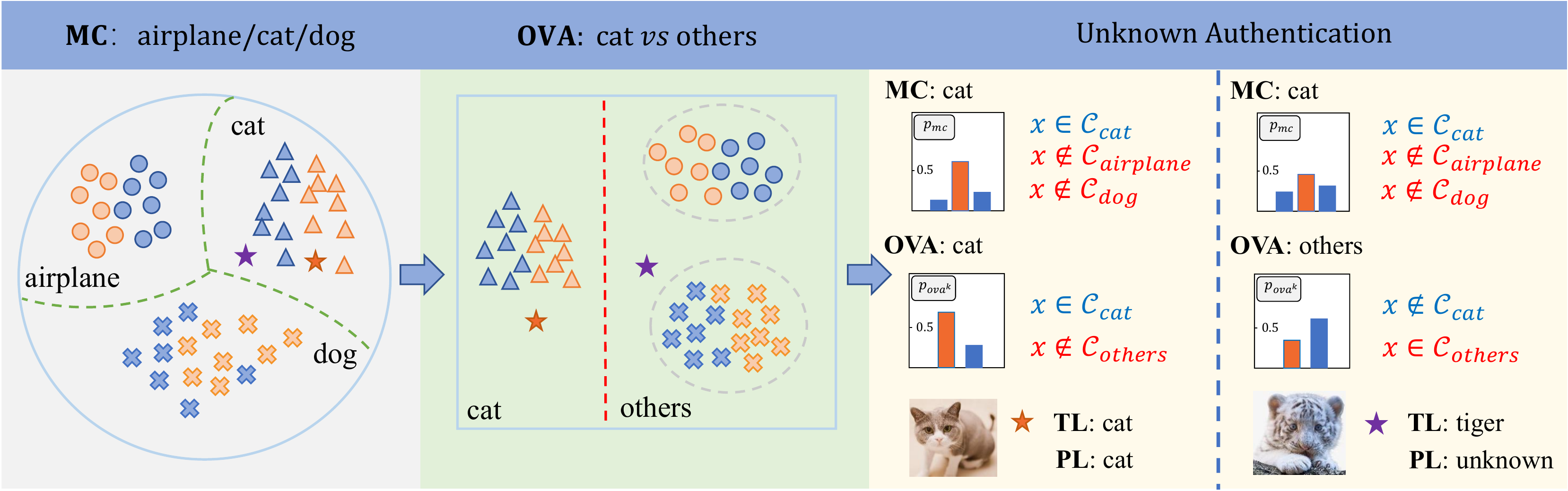}
    \caption{ Illustration of adaptive unknown authentication by prediction paradox from MC predictor and the corresponding OVA predictor, if the predictions for sample \emph{x} from MC and the OVA predictor are consistent, the predicted label (PL) is among the known source classes, otherwise, it is predicted as an ``unknown'' sample.}
    \label{fig1}
\end{figure}

In UACP, a composite classifier is designed with two types of predictors for classification and verification, respectively. The MC predictor classifies samples to one of the multiple source classes, and the corresponding binary OVA predictor further verifies if a sample belongs to the predicted class by MC predictor. Samples with paradoxical predictions are rejected as unknowns. An illustration is shown in Figure~\ref{fig1}, the sample with true label (TL) ``tiger'' denoted by \emph{purple star} is classified to ``cat'' by MC predictor, thus it is declined by the ``airplane" and ``dog". But meanwhile, the corresponding OVA predictor (cat vs others) gives a paradoxical prediction that it belongs to the ``others" rather than ``cat", then it is predicted as an ``unknown" sample not included in the source classes, since the output space of OVA predictor contains both known source and unknown classes in the negative ``others" category. At the same time, the sample with TL ``cat" denoted by \emph{orange star} has consistent predictions from the two predictors, thus it is classified to the known ``cat" class. Moreover, different from previous feature alignment for shared classes, implicit domain alignment is conducted in the output space by a domain-invariant classifier. Specifically, features are generated for both domains such that the classifier correctly classifies source samples, and captures the target structure as well. In this way, samples across domains share the same decision boundary, though with different feature distributions. The main contributions of this paper are organized as follows,
\begin{itemize}
\item {We propose adaptive unknown authentication by classifier paradox for UniDA, such that target unknowns are adaptively identified by prediction paradox from two types of predictors.}
\item We propose implicit domain alignment in the output space for UniDA, such that samples across domains share the same decision boundary, though exhibit feature discrepancy.
\item Empirical comparisons with state-of-arts validate the proposed UACP in both open-set and universal UDA settings.
\end{itemize}

\section{Related Works}\label{sec2}

In this section, we briefly review the related UDA methods, including closed-set UDA, OSDA, and UniDA methods in separate sub-sections.

\subsection{Unsupervised Domain Adaptation}\label{subsec21}

Closed-set UDA is the classical scenario in which source and target domains have distribution shift but consistent label sets. UDA approaches commonly attempt to reduce the distribution discrepancy for domain-invariant features across domains. The two main categories are statistical-based methods and adversarial-based methods. Statistical-based UDA methods directly minimize a discrepancy metric across domains, such as MMD (Tzeng et al., \citeyear{tzeng2014deep}), multi-kernel MMD (Long et al., \citeyear{long2015learning}), joint MMD (Long et al., \citeyear{long2017deep}), and correlation (Sun et al., \citeyear{sun2016deep}), etc. Adversarial-based methods maximize the domain confusion via adversarial learning between feature generator and domain discriminator, or between different classifiers (Bousmalis et al., \citeyear{bousmalis2016domain}; Ganin et al., \citeyear{ganin2016domain}; Saito et al., \citeyear{saito2018maximum}). Besides, some works also utilize learning strategies from other fields, such as curriculum learning (Choi et al., \citeyear{choi2019pseudo}), co-training (S. Wu et al., \citeyear{wu2019improving}), self-training (Yang et al., \citeyear{yang2021st3d}), and entropy regularization (X. Wu et al., \citeyear{wu2021entropy}), etc.

\subsection{Open-Set Domain Adaptation}\label{subsec22}

In OSDA, the target domain contains novel categories that are not observed in the source domain. Existing OSDA methods tackle this problem by first separating target unknowns from known data, and performing feature alignment on shared classes across domains. Panareda Busto \& Gall (\citeyear{panareda2017open}) assign the target samples to one of the known and unknown classes based on distance of target features to source centroids. Saito et al. (\citeyear{saito2018open}) adopt an adversarial framework for OSDA, and manually set a pre-defined threshold to either align target samples with source data or reject them as unknowns. Pan et al. (\citeyear{pan2020exploring}) present a self-ensemble method to exploit the category-agnostic clusters for unknown detection. Liu et al. (\citeyear{liu2019separate}) employ a coarse-to-fine weighting mechanism to progressively identify target unknowns. Bucci et al. (\citeyear{bucci2020effectiveness}) utilize rotation recognition with tailed adjustments to separate known and unknown target samples, and align distributions of shared classes across domains.

\subsection{Universal Domain Adaptation}\label{subsec23}

In UniDA, both domains may contain unshared or private classes, while no prior information about the target label set is provided. You et al. (\citeyear{you2019universal}) quantify sample-level transferability to distinguish shared and private classes in each domain. Later on, Saito et al. (\citeyear{saito2020universal}) apply neighborhood clustering and entropy separation to encourage known target samples close to source prototypes while away from unknown classes. Fu et al. (\citeyear{fu2020learning}) present calibrated multiple uncertainty to detect open classes more accurately. Li et al. (\citeyear{li2021domain}) utilize the intrinsic structure of target samples and provide a unified framework to deal with different sub-cases of UniDA. Saito \& Saenko (\citeyear{saito2021ovanet}) adopt the minimum inter-class distance in source domain as the threshold to identify unknowns in target. Some recent researches also study UniDA in different scenarios. Yu et al. (\citeyear{yu2021divergence}) adopt the divergence between two classifiers as sample confidence in noisy UniDA. Liang et al. (\citeyear{liang2021umad}) develop an informative consistency score based on two classifiers to help distinguish unknown samples in source-free UniDA.

Different from existing works (Yu et al., \citeyear{yu2021divergence}; Liang et al., \citeyear{liang2021umad}) that adopt the discrepancy of two same-structured classifiers for describing sample ``confidence", our proposed UACP exploits the prediction paradox between the two types of predictors (MC and OVA) to directly identify unknowns. Besides, OVANet (Saito \& Saenko, \citeyear{saito2021ovanet}) utilizes OVA classifiers for UniDA, in order to seek the minimum source inter-class distance as the unknown threshold, while UACP adopts OVA classifiers for adaptive unknown authentication.

\section{Methodology}\label{sec3}

In this section, we introduce UACP for UniDA by classifier paradox. We first revisit the problem setting of UniDA, and describe the network architecture of UACP. After that, we show individual components in UACP in detail.

\begin{figure}
    \centering
    \includegraphics[width=1.0\textwidth]{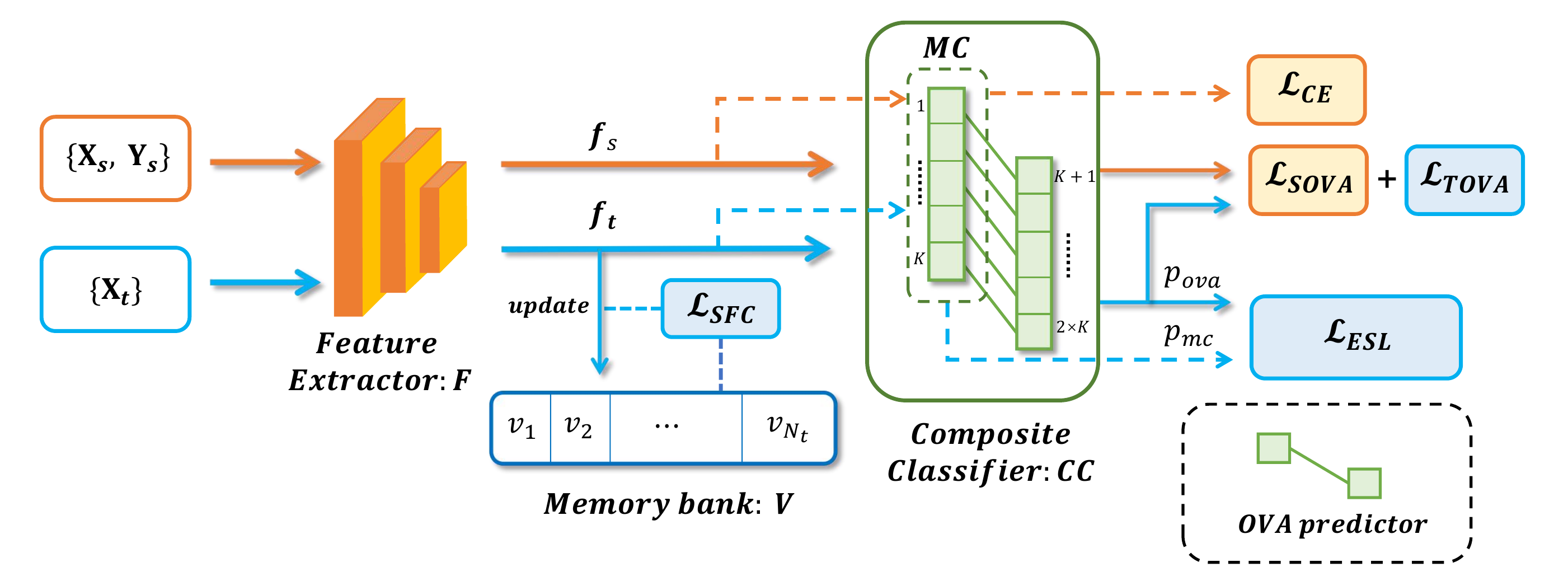}
    \caption{Overall framework of UACP, which includes a shared feature extractor $F$ for source and target domains, and a composite classifier with a MC predictor and binary OVA predictors. The MC and OVA predictors share the layer neurons, i.e., the first $K$ neurons construct the MC predictor, while the $k$-th and $(K+k)$-th neurons construct the OVA predictor for the $k$-th class.}
    \label{fig2}
\end{figure}
\subsection{Problem Setting and Network Architecture}\label{subsec31}

In UniDA, we are given a labeled source domain $\mathcal{D}_s={\{({x}_i^s,{y}_i^s)\}}_{i=1}^{N_s}$ with ${N}_s$ labeled source samples, and an unlabeled target domain $\mathcal{D}_t={\{({x}_i^t)\}}_{i=1}^{N_t}$ with ${N}_t$ unlabeled target samples, where ${x}_i^s$ and ${x}_i^t$  denote the source and target samples, respectively. ${y}_i^s\in{\{1,\cdots,K\}}$ is the class label for sample ${x}_i^s$, and $K$ is the number of source classes. Define $L_s$ and $L_t$ as the label sets of source and target domains, respectively. Then the class set shared across domains is denoted as $L_s\cap L_t$. $L_s-L_t$ and $L_t-L_s$ represent the source-private and target-private class sets, respectively. We mainly focus on the scenario with ${L_t-L_s}{\neq} {\emptyset}$, including both OSDA and UniDA, and the learning goal is to classify target samples into $\lvert{L_s}\cap{L_t}\rvert+1$ classes, that is, to classify known target samples to the shared source classes, while recognize unknown target samples in all target-private classes as well.

The architecture of UACP is given in Figure~\ref{fig2}. It contains two components: (i)~a feature extractor $F$, which outputs the $\mathcal \ell_2$ normalized feature vector, and (ii)~a composite classifier $CC$ composed of ${2}\times{K}$ neurons. In $CC$, a MC predictor w.r.t. the first $K$ neurons is adopted to classify target samples to one of the $K$ source classes. Further, totally $K$ OVA predictors are further adopted to verify the prediction from MC predictor, and the OVA predictor for the $k$-th class is built on the $k$-th and $(K+k)$-th neurons. The memory bank holds features for all target samples currently. For a target sample $x$, $p_{mc}(x)$ and $p_{ova}(x)$ denote the predictions from MC and the corresponding OVA predictor, respectively. 

\subsection{Composite Classifier with MC and OVA predictors}\label{subsec32}

In order to adaptively identify unknown target samples, a novel composite classifier with both MC and OVA predictors is designed in UACP. The MC predictor outputs a $K$-dimensional vector, in order to classify samples to one of the source classes. Further, there are also $K$ OVA predictors represented as $OVA=\{ova^1,ova^2,\cdots,ova^K\}$, each OVA predictor related to one source class verifies the prediction by MC predictor. Let $p_{mc}=\sigma(MC(F(x)))\in{\mathbb{R}^K}$ denote the probability output vector for sample $x$ by MC predictor, where $\sigma$ is the softmax function, each dimension $p_{mc}^k(x)$ describes the probability of $x$ to the $k$-th class. $p_{{ova}^k}=\sigma({ova}^k(F(x)))\in{\mathbb{R}^2}$ denotes the probability output for $x$ by the $k$-th OVA predictor, in which $p_{{ova}^k}^+$ and $p_{{ova}^k}^-$ are the probabilities of $x$ to the $k$-th (positive) and rest (negative) classes, respectively.

To obtain discriminative features among different categories, we minimize the cross-entropy loss with source supervision for MC predictor,
\begin{equation}
    \mathcal{L}_{CE}=\mathbb{E}_{(x_i^s,y_i^s)\in{\mathcal{D}_s}}\ell_{ce}(p_{mc}(x_i^s),y_i^s)\label{eq1}
\end{equation}
where $\ell_{ce}$ is the standard cross-entropy loss.

Due to the property that the OVA predictor does not enforce each sample to belong to only source classes, UACP adopts it to further verify the prediction by MC predictor. For each source class, an OVA predictor learns the decision boundary between the positive in-class and negative out-class categories, and the negative category actually includes both source and unknown classes. At the same time, UACP learns discriminative features among different categories by maximizing the distance between similar categories (Padhy et al., \citeyear{padhy2020revisiting}; Saito \& Saenko, \citeyear{saito2021ovanet}). Specifically, for each source sample $x_i^s$, the discrepancy between $ova^{y_i^s}$ and OVA predictor w.r.t. the closest negative class is further maximized,
\begin{equation}
    \mathcal{L}_{SOVA}=\mathbb{E}_{(x_i^s,y_i^s)\in{\mathcal{D}_s}}\left \{-log(p_{{ova}^{y_i^s}}^+(x_i^s))+\mathop{max}\limits_{j\neq{y_i^s}}[log(p_{{ova}^j}^+(x_i^s))]\right \}\label{eq2}
\end{equation}
where $j$ represents the closest negative class for $x_i^s$. Through minimizing the above loss, OVA predictors can not only identify in-class and out-class samples, but also separate similar classes with source supervision.

\subsection{Prediction Paradox for Target Unknown Authentication}\label{subsec33}

Since some target categories are absent in source domain, it is difficult to make accurate predictions for target samples directly with source classifier, especially for target unknowns. To the end, UACP adopts classifier paradox for adaptive unknown authentication, including a MC predictor to classify samples to one of the multiple source classes, and a corresponding OVA predictor to further verify whether the sample belongs to the predicted class or not. If the OVA predictor affirms the prediction by MC predictor, which means MC and OVA have consistent predictions, then it is confident to predict the sample to a known class in source domain. Otherwise, if there is verification failure by OVA predictor, or the MC and OVA predictors have paradoxical predictions, then the sample tends to belong to unknown class. Finally, for each sample $x_i$, let $k=argmax(p_{mc}(x_i))$ denote the predicted class by MC predictor, then
\begin{equation}
    x_i\in{\left\{
    \begin{aligned}
    \emph{C}_k,~~~~~~~~&   &   if~p_{{ova}^{k}}^+(x_i)\geq p_{{ova}^{k}}^-(x_i)\\
    \emph{C}_{unknown},&   &   if~p_{{ova}^{k}}^+(x_i)< p_{{ova}^{k}}^-(x_i)\\
\end{aligned}
\right.}\label{eq3}
\end{equation}
where $\emph{C}_k$ and $\emph{C}_{unknown}$ denote the \emph{k}-th known class and unknown class, respectively.

Further, we adopt an entropy-strengthened loss over target samples for MC predictor, in order to strengthen the consistency between MC and OVA predictors,  and capture the low-density separation for target samples in classifier learning as well. Specifically, for a known target sample with MC prediction affirmed by OVA predictor, we further constrain a sharper probability distribution or more confident prediction in MC predictor, while for an unknown sample with prediction paradox, a more uniform distribution or less confident prediction is further expected. Finally, for each target sample $x_i^t$, the entropy-strengthened loss is expressed as,
\begin{equation}
    \mathcal{L}_{ESL}(x_i^t)={\left\{
    \begin{aligned}
    -p_{mc}(x_i^t)log(p_{mc}(x_i^t)),&   &   if~p_{{ova}^{k}}^+(x_i^t)> p_{{ova}^{k}}^-(x_i^t)+m\\
    p_{mc}(x_i^t)log(p_{mc}(x_i^t)),&   &   if~p_{{ova}^{k}}^+(x_i^t)< p_{{ova}^{k}}^-(x_i^t)-m\\
    0,~~~~~~~~~~~~&   &   otherwise~~~~~~~~~~~  
\end{aligned}
\right.}\label{eq4}
\end{equation}
and
\begin{equation}
    \mathcal{L}_{ESL}=\mathbb{E}_{(x_i^t)\in\mathcal{D}_t}\mathcal{L}_{ESL}(x_i^t)\label{eq5}
\end{equation}
where $k$ is the predicted class for sample $x_i^t$ by MC predictor, and $m$ is the margin for selecting confident known and unknown samples. In particular, with our special design of composite classifier, the predictions from MC and OVA tend to be consistent due to their partial-shared parameters. It is noted that we adopt $m$ to conduct constraint on confident target samples, in order to exclude the incorrect predictions from MC predictor.

\subsection{Domain-invariant Classifier for Implicit Domain Alignment}\label{subsec34}

Previous UniDA methods commonly reduce the domain shift by feature alignment for shared classes across domains, while a mis-identification of target unknowns further incurs mis-alignment of these classes. In UACP, an implicit domain alignment is conducted directly in the output space. Specifically, a domain-invariant classifier is trained such that samples across domains share the same decision boundary, though with exhibit different feature distributions.

First, the source supervision is adopted for both feature extractor and classifier learning in Sect.~\ref{subsec32}. Due to the lack of target ground-truth, we further propose to leverage the self-supervised knowledge from target data. Our idea is that nearby samples should be close to each other in feature space, so as to generate well-clustered features for target data. A memory bank is utilized as $V=\{v_1,v_2,\cdots, v_{N_t}\}$, where $v_i$ is the stored features vector for $x_i$, and it is updated with the mini-batch features in each iteration. Then the similarity between feature $f_i=F(x_i)$ and stored feature $v_j$ with $i\neq{j}$ is calculated as,
\begin{equation}
    p_{i,j}=\frac{exp(v_j^\top f_i)/\tau}{\sum_{r=1,r\neq{i}}^{N_t}exp(v_j^\top f_i)/\tau}\label{eq6}
\end{equation}
where temperature $\tau$ determines the level of concentration (Hinton et al., ~\citeyear{hinton2015distilling}). $p_{i,j}$ actually describes the probability that feature $f_i$ is a neighbor of $v_j$. To enforce samples be close to its nearby neighbors, a self-supervised feature clustering loss is adopted for target samples as,
\begin{equation}
    \mathcal{L}_{SFC}=-\mathbb{E}_{(x_i^t)\in \mathcal{D}_t}\sum_{j=1,j\neq{i}}^{N_t}p_{i,j}log(p_{i,j})\label{eq7}
\end{equation}
Minimizing the above loss actually minimizes the entropy of each target sample’s similarity distribution to other target samples, thus helps gather similar target samples together to form compact clusters, and separate target samples from different clusters in the feature space.

Further, the low-density separation for target data has already been  enforced over MC predictor in Eq.~(\ref{eq5}). Besides, it is also conducted on OVA predictors to seek a domain-invariant classifier across domains. Specifically, we perform entropy minimization (Saito et al., \citeyear{saito2019semi}) for target samples over each OVA predictor by,
\begin{equation}
    \mathcal{L}_{TOVA}=-\mathbb{E}_{(x_i^t)\in \mathcal{D}_t}\sum_{k=1}^{K}\{p_{{ova}^{k}}^+(x_i^t)log(p_{{ova}^{k}}^+(x_i^t))+p_{{ova}^{k}}^-(x_i^t)log(p_{{ova}^{k}}^-(x_i^t))\}\label{eq8}
\end{equation}
The above loss is minimized to increase the prediction confidence of OVA predictors. In this way, the shared classes across domains are implicitly aligned, while the target unknowns are kept away from the known classes.
\begin{algorithm}[t]
    \caption{UACP}
    \hspace*{0.02in} {\bf Input:} Source domain $\mathcal{D}_s=\{(x_i^s,y_i^s)\}_{i=1}^{N_s}$; Target domain $\mathcal{D}_t=\{(x_i^t)\}_{i=1}^{N_t}$;\\
    \hspace*{0.02in} Pre-trained feature extractor $F$ parameterized by $\theta_F$;\\
    \hspace*{0.02in} Randomly initialized composite classifier $CC$ parameterized by $\theta_{CC}$;\\
    \hspace*{0.02in} Number of iterations $T$ in an epoch; Number of epochs $E$.\\
    \hspace*{0.02in} {\bf Output:} Optimal parameters $\theta_{F}$, $\theta_{CC}.$
        \begin{algorithmic}[1]
        \For{epoch=$1$ to $E$}
            \State Initialize memory bank $V$ with target features.
            \For{iteration=$1$ to $T$}
                \State Sample a batch ($\mathcal{S}$) from $\mathcal{D}_s$ and ($\mathcal{T}$) from $\mathcal{D}_t.$
                \State Obtain extracted features $f_s=F(\mathcal{S})$ and $f_t=F(\mathcal{T}).$
                \State Update features at the current positions in $V$ with $f_t$.
                \State Compute the similarities between $f_t$ and features in $V$ using Eq.~(\ref{eq6}).
                \State Obtain probability outputs of $CC$: $p_{mc}$ and $p_{ova}$.
                \State Compute $\mathcal{L}_{CE}(\mathcal{S})$ and $\mathcal{L}_{SOVA}(\mathcal{S})$ using source data.
                \State Compute $\mathcal{L}_{SFC}(\mathcal{T})$, $\mathcal{L}_{TOVA}(\mathcal{T})$, and $\mathcal{L}_{ESL}(\mathcal{T})$ using target data.
                \State Update $\theta_F$ and $\theta_{CC}$ by minimizing the overall loss in Eq. (\ref{eq9}).
            \EndFor
        \EndFor
    \end{algorithmic}
    \label{algo1}
\end{algorithm}
\subsection{Overall Training Objective for UACP}\label{subsec35}

The final learning objective of UACP can be formulated as,
\begin{equation}
    % \mathop{min}\limits_{\theta_F,\theta_{CC}}&\mathcal{L}_{SRC}+\mathcal{L}_{TGT}
    \mathop{min}\limits_{\theta_F,\theta_{CC}}(\mathcal{L}_{CE}+\mathcal{L}_{SOVA})+(\alpha\cdot\mathcal{L}_{SFC}+\beta\cdot\mathcal{L}_{TOVA}+\gamma\cdot\mathcal{L}_{ESL}) \label{eq9}
\end{equation}
where $\alpha$, $\beta$, and $\gamma$ control the trade-off for each component in Eq.~(\ref{eq9}). In each iteration, the memory bank updates a batch target features and the network updates parameters $\theta_F$ and $\theta_{CC}$. The algorithm description for UACP is given in Algorithm~\ref{algo1}.

\section{Experiments}\label{sec4}
We validate the proposed UACP mainly with two adaptation settings, open-set domain adaptation and universal domain adaptation.

\subsection{Setup}\label{subsec41}
\subsubsection{Datasets}\label{subsubsec411}
We perform experiments on four popular benchmark datasets in UDA: Office-31 (Saenko et al., \citeyear{saenko2010adapting}), Office-Home (Venkateswara et al., \citeyear{venkateswara2017deep}), VisDA (Peng et al., \citeyear{peng2017visda}), and DomainNet (Peng et al., \citeyear{peng2019moment}). Office-31 consists of 31 categories in 3 domains, i.e., DSLR(D), Amazon(A) and Webcam(W), with totally 4652 images. Office-Home contains 15500 images from 4 different domains, i.e., Artistic images (Ar), Clipart images (Cl), Product images (Pr), and Real-World images (Re). VisDA is a more challenging dataset, which consists of 15K source synthetic images and 5K target natural images. DomainNet contains 345 classes and 6 domains, and following Fu et al. (\citeyear{fu2020learning}), we use 3 domains Painting(P), Real(R), and Sketch(S). Similar to Li et al. (\citeyear{li2021domain}), we split classes into three parts: common classes across domains $\lvert{L_s}\cap{L_t}\rvert$, source-private classes $\lvert{L_s}-{L_t}\rvert$, and target-private classes $\lvert{L_t}-{L_s}\rvert$. The details of category splits in the four datasets are shown in Table~\ref{tab1}.

\begin{table}[h]
\centering
\begin{center}
\begin{minipage}{0.6\textwidth}
\caption{The division on label sets in each setting}\label{tab1}%
\resizebox{\textwidth}{!}{
\begin{tabular}{@{}llccc@{}}
\toprule
Tasks & Datasets  & $\lvert{L_s \cap L_t}\rvert$ & $\lvert{L_s-L_t}\rvert$ & $\lvert{L_t-L_s}\rvert$\\
\midrule
\multirow{3}{*}{OSDA}
 & Office-31     & 10  & 0  & 11  \\
 & Office-Home   & 25  & 0  & 40  \\
 & VisDA         & 6   & 0  & 6  \\
\midrule
\multirow{4}{*}{UniDA}
 & Office-31    & 10  & 10 & 11  \\
 & Office-Home  & 10  & 5  & 50  \\
 & VisDA        & 6   & 3  & 3  \\
 & DomainNet    &150  & 50 & 145\\
\botrule
\end{tabular}
}
\end{minipage}
\end{center}
\end{table}

\subsubsection{Evaluation Metrics}

To better evaluate the performance of UACP under both OSDA and UniDA scenarios, we utilize the $HOS$ metric (Bucci et al., \citeyear{bucci2020effectiveness}) defined as the harmonic mean of average per-class accuracies over known and unknown samples, denoted by $Acc_{kn}$ and $Acc_{unk}$, respectively. $HOS$ is formulated as,
\begin{equation}
    HOS=2\times{\frac{Acc_{kn}\times{Acc}_{unk}}{Acc_{kn}+Acc_{unk}}}
\end{equation}
 It fairly considers performance on known and unknown data. Besides, instance-wise accuracy on known classes ($Acc$) and area under the ROC curve ($AUC$) are also adopted in Sect. \ref{subsub434}, following the standard protocol of unknown detection (Hendrycks \& Gimpel, \citeyear{hendrycks2016baseline}).

\subsubsection{Implementation Details}\label{subsubsec414}

We perform UACP in Pytorch (Paszke et al., \citeyear{paszke2017automatic}) framework. For fair comparisons, we implement our network based on ResNet-50 (He et al., \citeyear{he2016deep}) pre-trained on ImageNet (Russakovsky et al., \citeyear{russakovsky2015imagenet}). Following Saito et al. (\citeyear{saito2019semi}), the last linear layer is replaced by a new linear classification layer. The learning rates for the new linear layer and finetuned layers with inverse scheduling are set to 0.01 and 0.001, respectively. We exploit a mini-batch SGD optimizer with momentum 0.9 and weight decay 0.0005 in all experiments. The value of temperature $\tau$ is set to 0.05 following Ranjan et al. (\citeyear{ranjan2017l2}). In UACP, trade-off parameters $\alpha$, $\beta$, and $\gamma$ are fixed, i.e., $\alpha$=$\gamma$=0.05, and $\beta$=0.1. $m$ is set to 0.4 for Office-31 and Office-Home, while 0.5 for VisDA and DomainNet.

\subsection{Reults}\label{subsec42}
In this section, we evaluate UACP by comparing with the state-of-the-arts. The bolded value in each column indicates the best performance of all methods.

\subsubsection{Open-Set Domain Adaptation}\label{subsubsec421}

We perform comparisons under OSDA scenario over Office-31, Office-Home and VisDA datasets. There are 6 tasks for Office-31 and 12 tasks for Office-Home. We compare with OSDA methods OSBP (Saito et al., \citeyear{saito2018open}), STA (Liu et al., \citeyear{liu2019separate}) and ROS (Bucci et al., \citeyear{bucci2020effectiveness}), as well as UniDA methods UAN (You et al., \citeyear{you2019universal}), DANCE (Saito et al., \citeyear{saito2020universal}), CMU (Fu et al., \citeyear{fu2020learning}), DCC (Li et al., \citeyear{li2021domain}) and OVANet (Saito \& Saenko, \citeyear{saito2021ovanet}).

The results over Office-31 and VisDA datasets are reported in Table~\ref{tab2}, and Table~\ref{tab3} records the performance over Office-Home dataset. UACP achieves the best performance on 5 of the 6 tasks on Office-31. On average, it outperforms previous state-of-the-art method OVANet by 3.0\%. As for VisDA dataset, UACP significantly outperforms OVANet by 16.3\%, and outperforms the second-best method DCC by 1.7\%. Note that DCC takes the prior of OSDA and UniDA settings into consideration, while our UACP has no prior knowledge of private classes. From Table~\ref{tab3}, UACP achieves the best results on 7 of 12 tasks. On average, UACP surpasses all OSDA and UniDA methods. Notably, OSDA methods are designed for specific open-set scenario that relies on prior knowledge, while UACP is able to adapt to more general scenarios.
\begin{table}[tb]
    \centering
    \begin{center}
        \begin{minipage}{0.8\textwidth}
        \caption{$HOS$ (\%) on Office-31 and VisDA for OSDA}\label{tab2}
        \resizebox{\textwidth}{!}{
\begin{tabular}{@{}lccccccc|c@{}}
\toprule
 Method & A$\to$D & A$\to$W & D$\to$A  & D$\to$W & W$\to$D & W$\to$A & Avg & VisDA\\
\midrule
OSBP   & 82.4    & 82.7    & 75.1     & 97.2    & 91.1    & 73.7    & 83.7 & 46.9 \\
STA    & 75.0    & 75.9    & 73.2     & 69.8    & 75.2    & 66.1    & 72.5 & - \\
ROS    & 82.4    & 82.1    & 77.9     & 96.0    & $\bf{99.7}$    & 77.2    & 85.9 & 50.1 \\
UAN    & 54.2    & 57.4    & 73.7     & 75.2    & 67.6    & 59.8    & 64.6 & 50.8 \\
DANCE  & 82.0    & 74.7    & 68.0     & 82.1    & 82.5    & 52.5    & 73.6 & 59.7 \\
CMU    & 71.6    & 70.5    & 80.2     & 81.2    & 70.8    & 70.8    & 74.2 & 24.1 \\
DCC    & 85.5    & 87.1    & 85.5     & 91.2    & 87.1    & 84.4    & 86.8 & 70.7 \\
OVANet & 90.5    & 88.3    & 86.7     & 98.2    & 98.4    & 88.3    & 91.7 & 56.1 \\
UACP   &$\bf{92.3}$ & $\bf{92.7}$ & $\bf{93.3}$ & $\bf{98.8}$ & 99.4 & $\bf{91.6}$ & $\bf{94.7}$ & $\bf{72.4}$ \\
\botrule
\end{tabular}}
        \end{minipage}
    \end{center}
    
\end{table}

\begin{table*}[tb]
    % \centering
    \begin{minipage}{\textwidth}
    \caption{$HOS$ (\%) on Office-Home for OSDA}\label{tab3}
    
    \setlength{\tabcolsep}{0.05cm}{
    \resizebox{\textwidth}{!}{
    \begin{tabular}{lccccccccccccc}
         \toprule
        %  Dataset & \multicolumn{13}{c}{Office-Home} \\
        %  \midrule
         Method&Ar$\to$Cl&Al$\to$Pr&Ar$\to$Re&Cl$\to$Ar&Cl$\to$Pr&Cl$\to$Re&Pr$\to$Al&Pr$\to$Cl&Pr$\to$Re&Re$\to$Al&Re$\to$Cl&Re$\to$Pr&Avg \\
         \midrule
         OSBP   & 55.1 & 65.2 & 72.9 & $\bf{64.3}$ & 64.7 & 70.6 & $\bf{63.2}$ & 53.2 & 73.9 & 66.7 & 54.5 & 72.3 & 64.7 \\
         STA    & 55.8 & 54.0 & 68.3 & 57.4 & 60.4 & 66.8 & 61.9 & 53.2 & 69.5 & 67.1 & 54.5 & 64.5 & 61.1 \\
         ROS    & 60.1 & 69.3 & $\bf{76.5}$ & 58.9 & 65.2 & 68.6 & 60.6 & $\bf{56.3}$ & 74.4 & 68.8 & $\bf{60.4}$ & $\bf{75.7}$ & 66.2 \\
         UAN    & 34.7 & 22.4 & 9.4  & 38.9 & 22.9 & 21.8 & 47.4 & 39.7 & 30.9 & 34.4 & 35.8 & 22.0 & 30.0 \\
         DANCE  & 6.5  & 9.0  & 9.9  & 20.4 & 10.4 & 9.2  & 28.4 & 12.8 & 12.6 & 14.2 & 7.9  & 13.2 & 12.9 \\
         CMU    & 55.0 & 57.0 & 59.0 & 59.3 & 58.2 & 60.6 & 59.2 & 51.3 & 61.2 & 61.9 & 53.5 & 55.3 & 57.6 \\
         DCC    & 56.1 & 67.5 & 66.7 & 49.6 & 66.5 & 64.0 & 55.8 & 53.0 & 70.5 & 61.6 & 57.2 & 71.9 & 61.7 \\
         OVANet & 57.5 & 66.0 & 69.3 & 61.4 & 64.5 & 67.6 & 58.2 & 51.7 & 69.1 & 67.8 & 58.1 & 67.9 & 63.3 \\
         % temp
         UACP   & $\bf{60.3}$ & $\bf{69.5}$ & 73.8 & $\bf{64.3}$ & $\bf{67.1}$ & $\bf{71.9}$ & 62.8 & 54.3 & $\bf{74.8}$ & $\bf{69.9}$ & 58.5 & 70.8 & $\bf{66.5}$ \\
         \botrule
    \end{tabular}
    }
    }
    \end{minipage}
\end{table*}

\subsubsection{Universal Domain Adaptation}\label{subsubsec422}

As for UniDA, we perform UACP on Office-31, Office-Home, VisDA and DomainNet datasets. We compare UACP with OSDA methods, including OSBP (Saito et al., \citeyear{saito2018open}) and ROS (Bucci et al., \citeyear{bucci2020effectiveness}), and UniDA methods, including UAN (You et al., \citeyear{you2019universal}), DANCE (Saito et al., \citeyear{saito2020universal}), CMU (Fu et al., \citeyear{fu2020learning}), DCC (Li et al., \citeyear{li2021domain}), and OVANet (Saito \& Saenko, \citeyear{saito2021ovanet}).

Table~\ref{tab4} shows the results over Office31 dataset under UniDA scenario. The proposed UACP outperforms all the baselines on 5 of the 6 tasks, yielding 4.5\% improvement on average over previous state-of-the-art OVANet. The results on DomainNet and VisDA are recorded in Table~\ref{tab5}, and performance on Office-Home is reported in Table~\ref{tab6}. UACP achieves the best performance on two challenging benchmarks, i.e.,  60.1\% on VisDA and 74.7\% on Office-Home, which has improvements of 7.0\% and 2.9\% over the second-best method, respectively. For DomainNet dataset, UACP outperforms the baselines on 2 of 6 tasks, and achieves second-best $HOS$ of 50.3\%, which is actually comparable to the best one. In terms of the comparison results with both OSDA and UniDA methods in different settings, it can be observed that our proposed UACP is able to properly tackle different levels of category shifts, thus performs well in different UDA scenarios.
\begin{table}[tb]
    \centering
    \begin{center}
        \begin{minipage}{0.78\textwidth}
        \caption{$HOS$ (\%) on Office-31 for UniDA}\label{tab4}
        \resizebox{\textwidth}{!}{
\begin{tabular}{@{}lccccccc@{}}
         \toprule
         Method & A$\to$D & A$\to$W & D$\to$A  & D$\to$W & W$\to$D & W$\to$A & Avg \\
         \midrule
         OSBP   & 51.1    & 50.2    & 49.8     & 55.5    & 57.2    & 50.2    & 52.3  \\
         ROS    & 71.4    & 71.3    & 81.0     & 94.6    & 95.3    & 79.2    & 82.1 \\
         UAN    & 59.7    & 58.6    & 60.1     & 70.6    & 71.4    & 60.3    & 63.5 \\
         DANCE  & 78.6    & 71.5    & 79.9     & 91.4    & 87.9    & 72.2    & 80.3  \\
         CMU    & 68.1    & 67.3    & 71.4     & 79.3    & 80.4    & 72.2    & 73.1 \\
         DCC    & 88.5    & 78.5    & 70.2     & 79.3    & 88.6    & 75.9    & 80.2 \\
         OVANet & 85.8    & 79.4    & 80.1     & $\bf{95.4}$    & 94.3    & 84.0    & 86.5 \\
         UACP   &$\bf{89.2}$ & $\bf{85.8}$ & $\bf{89.7}$ & 94.2 & $\bf{97.6}$ & $\bf{89.4}$ & $\bf{91.0}$\\
         \botrule
\end{tabular}}
        \end{minipage}
    \end{center}
\end{table}
\begin{table}[tb]
    \centering
    \begin{center}
        \begin{minipage}{0.8\textwidth}
        \caption{$HOS$ (\%) on DomainNet and VisDA for UniDA}\label{tab5}
        \resizebox{\textwidth}{!}{
\begin{tabular}{@{}lccccccc|c@{}}
         \toprule
         Method & P$\to$R & P$\to$S & R$\to$P  & R$\to$S & S$\to$P & S$\to$R & Avg & VisDA\\
         \midrule
         OSBP   & 52.2    & 35.0    & 46.5     & 35.8    & 38.6    & 52.1    & 43.4 & 37.7 \\
         ROS    & 20.5    & 30.0    & 36.9     & 28.7    & 19.9    & 23.2    & 26.5 & 30.3 \\
         UAN    & 41.9    & 39.1    & 43.6     & 38.7    & 38.9    & 43.7    & 41.0 & 30.5 \\
         DANCE  & 21.0    & 37.0    & 47.3     & $\bf{46.7}$    & 27.7    & 21.0    & 33.5 & 4.4 \\
         CMU    & 50.8    & 45.1    & $\bf{52.2}$     & 45.6    & 44.8    & 51.0    & 48.3 & 34.6 \\
         DCC    & $\bf{56.9}$    & 43.7    & 50.3     & 43.3    & 44.9    & 56.2    & 49.2 & 43.0 \\
         OVANet & 56.0    & 47.1    & 51.7     & 44.9    & 47.4    & $\bf{57.2}$    & $\bf{50.7}$ & 53.1 \\
         UACP   & 54.8    & $\bf{47.7}$    & 50.9     & 43.7     & $\bf{48.5}$    & 56.4    & 50.3 & $\bf{60.1}$ \\
         \botrule
\end{tabular}}
        \end{minipage}
    \end{center}
\end{table}
\begin{table*}[tb]
    % \centering
    \begin{minipage}{\textwidth}
    \caption{$HOS$ (\%) on Office-Home for UniDA}
        \setlength{\tabcolsep}{0.05cm}{
    \resizebox{\textwidth}{!}{
    \begin{tabular}{lccccccccccccc}
         \toprule
        %  Dataset & \multicolumn{13}{c}{Office-Home} \\
        %  \midrule
         Method&Ar$\to$Cl&Al$\to$Pr&Ar$\to$Re&Cl$\to$Ar&Cl$\to$Pr&Cl$\to$Re&Pr$\to$Al&Pr$\to$Cl&Pr$\to$Re&Re$\to$Al&Re$\to$Cl&Re$\to$Pr&Avg \\
         \midrule
         OSBP   & 39.6 & 45.1 & 46.2 & 45.7 & 45.2 & 46.8 & 45.3 & 40.5 & 45.8 & 45.1 & 41.5 & 46.9 & 44.5 \\
         ROS    & 50.4 & 77.7 & $\bf{85.3}$ & 62.1 & 71.0 & 76.4 & 68.8 & 52.4 & 83.2 & 71.6 & 57.8 & 79.2 & 70.0 \\
         UAN    & 51.6 & 51.7 & 54.3 & 61.7 & 57.6 & 61.9 & 50.4 & 47.6 & 61.5 & 62.9 & 52.6 & 65.2 & 56.6 \\
         DANCE  & 34.1 & 23.9 & 38.3 & 46.7 & 21.6 & 35.4 & 58.2 & 47.5 & 39.4 & 32.8 & 38.3 & 43.1 & 38.3 \\
         CMU    & 56.0 & 56.9 & 59.2 & 66.9 & 64.3 & 67.8 & 54.7 & 51.1 & 66.4 & 68.2 & 57.9 & 69.7 & 61.6 \\
         DCC    & 57.9 & 54.0 & 58.0 & $\bf{74.6}$ & 70.6 & 77.5 & 64.3 & $\bf{73.6}$ & 74.9 & $\bf{80.9}$ & $\bf{75.1}$ & 80.3 & 70.1 \\
         OVANet & 62.8 & 75.6 & 78.6 & 70.7 & 68.8 & 75.0 & 71.3 & 58.6 & 80.5 & 76.1 & 64.1 & 78.9 & 71.8 \\
         UACP   & $\bf{64.1}$ & $\bf{78.5}$ & 82.1 & 72.1 & $\bf{73.9}$ & $\bf{79.8}$ & $\bf{78.1}$ & 60.6 & $\bf{83.3}$ & 79.4 & 62.7 & $\bf{81.6}$ & $\bf{74.7}$ \\
         \botrule
    \end{tabular}
    }}
    \label{tab6}
    \end{minipage}
\end{table*}

\subsection{Analysis}\label{subsec43}

In this section, more analyses are provided to further investigate the effectiveness of UACP.

\subsubsection{Varying the number of unknown classes}\label{subsubsec431}
\begin{figure}
    \centering

	\begin{minipage}{0.4\linewidth}\label{fig3a}
		\centerline{\includegraphics[width=\textwidth]{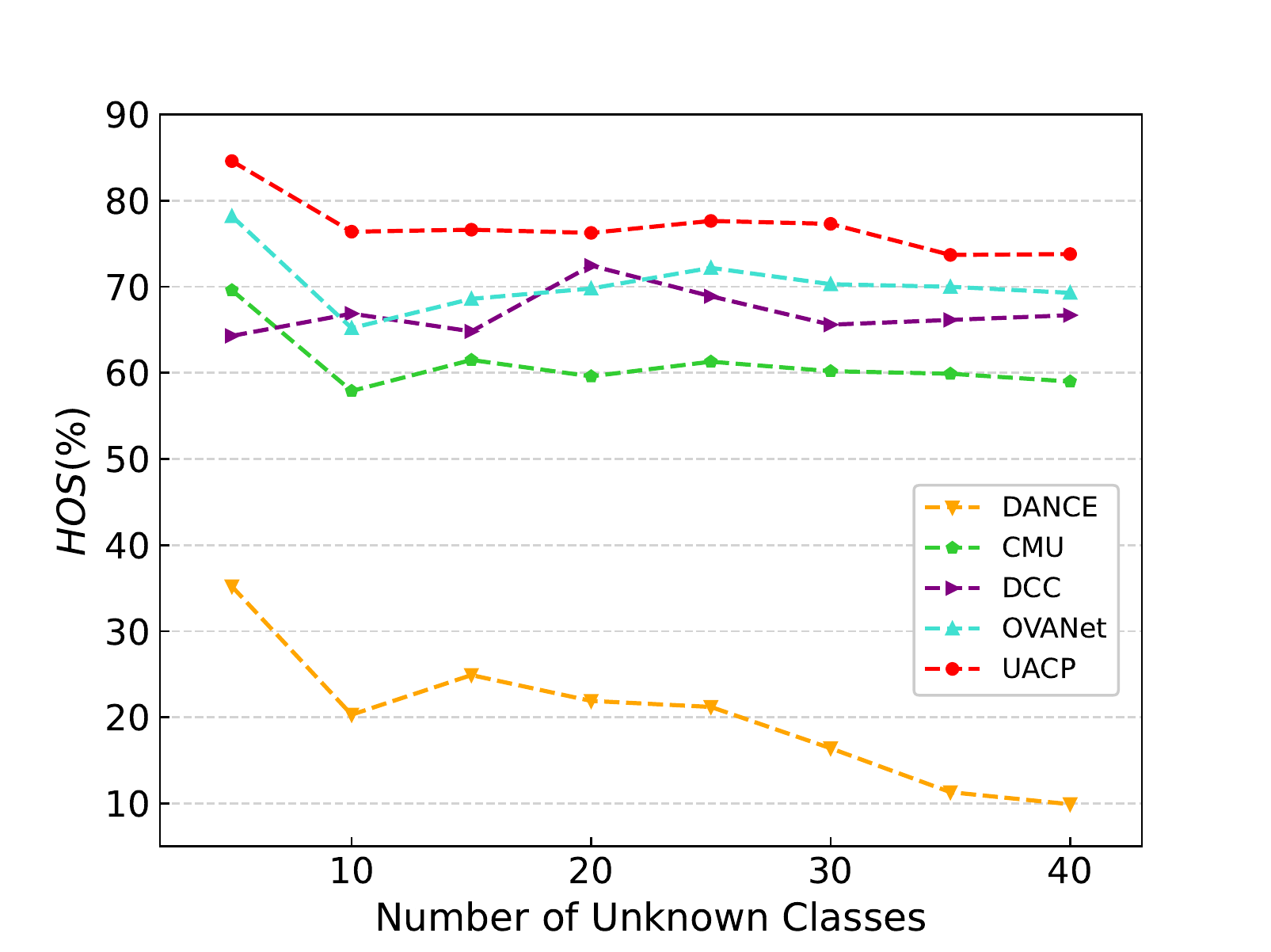}}
		\centerline{(a) Ar$\to$Re (OSDA)}
	\end{minipage}
	\begin{minipage}{0.4\linewidth}\label{fig3b}
		\centerline{\includegraphics[width=\textwidth]{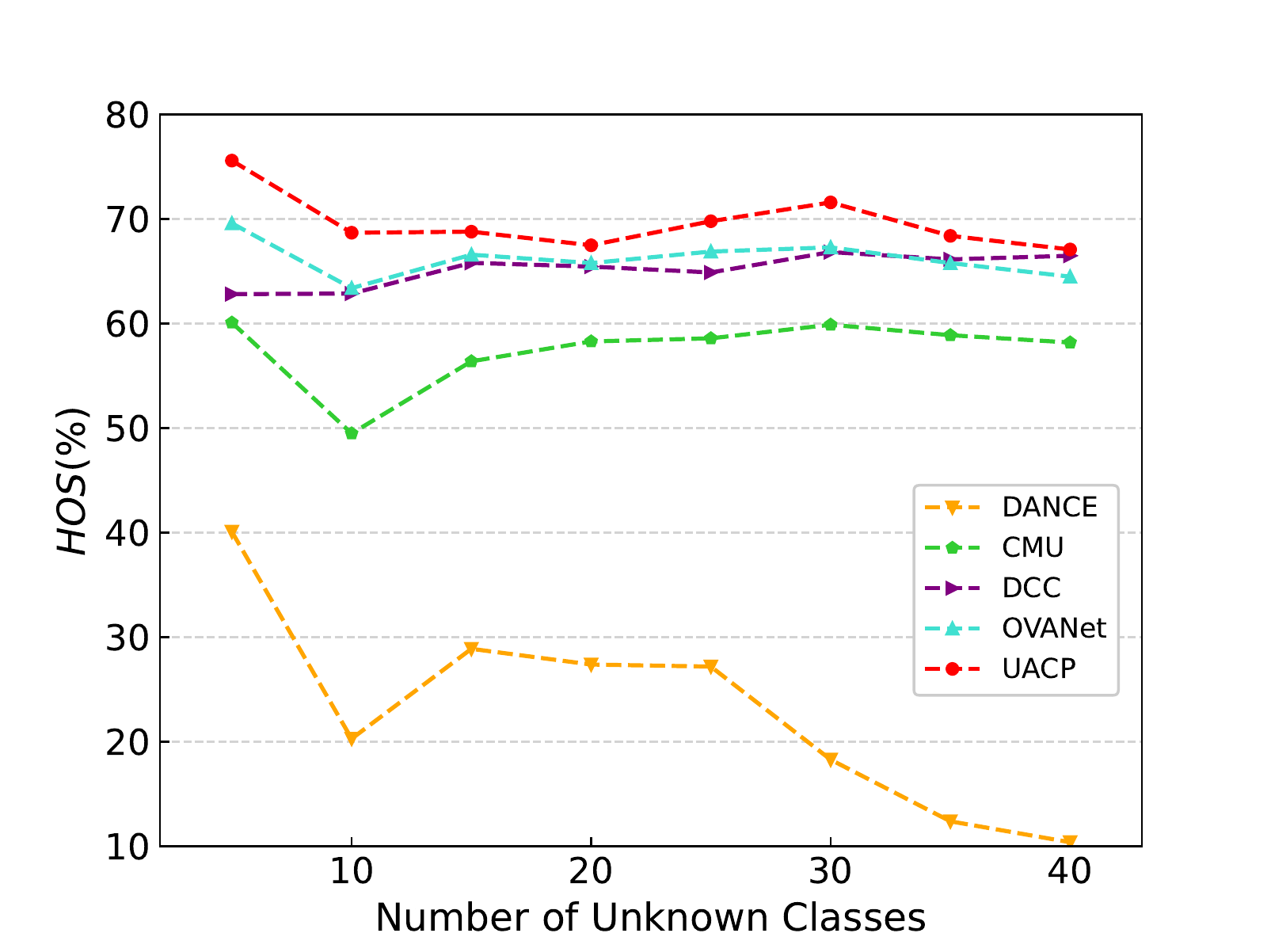}}
		\centerline{(b) Cl$\to$Pr (OSDA)}
	\end{minipage}
	\begin{minipage}{0.4\linewidth}\label{fig3c}
		\centerline{\includegraphics[width=\textwidth]{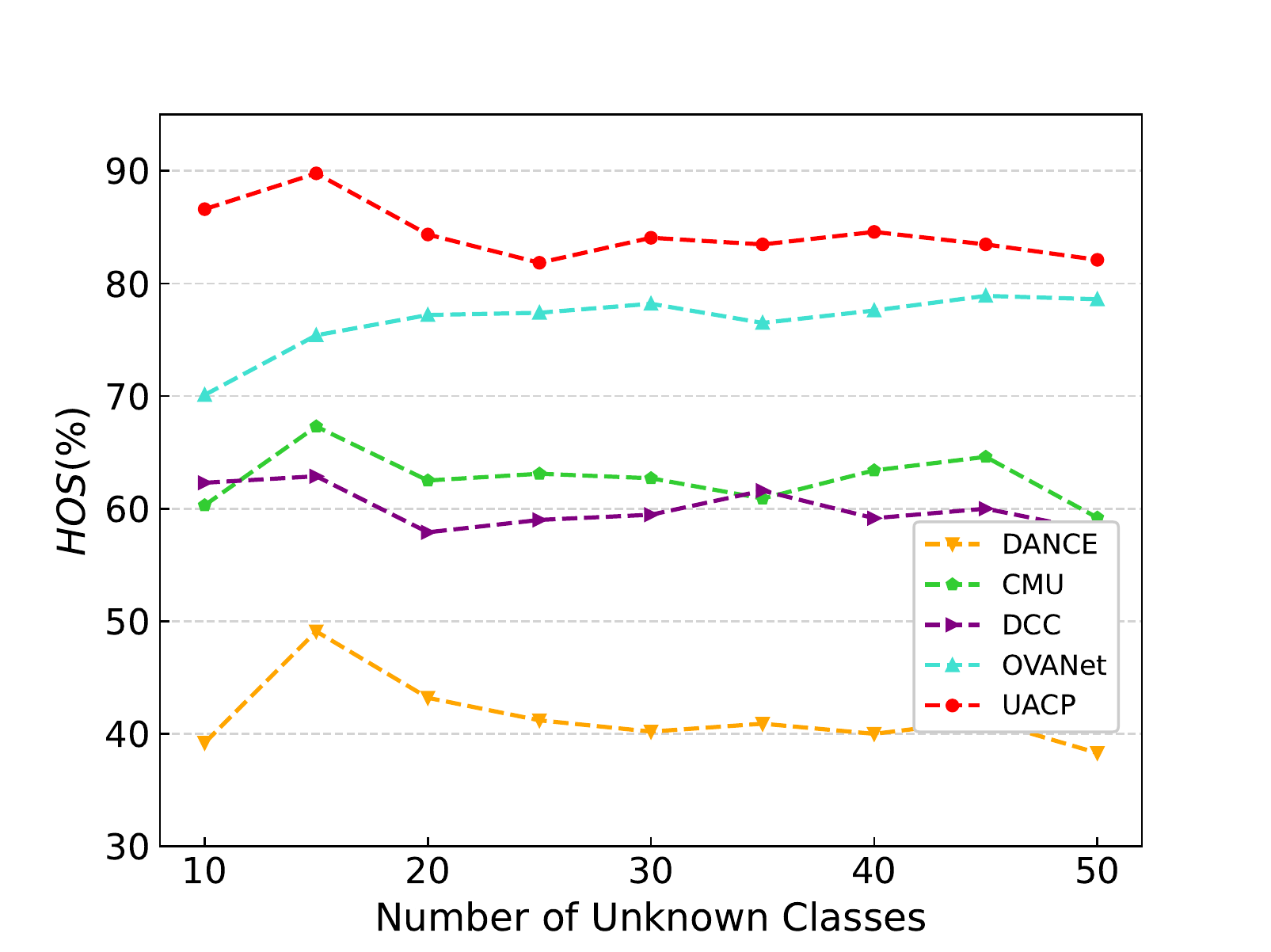}}
		\centerline{(c) Ar$\to$Re (UniDA)}
	\end{minipage}
	\begin{minipage}{0.4\linewidth}\label{fig3d}
		\centerline{\includegraphics[width=\textwidth]{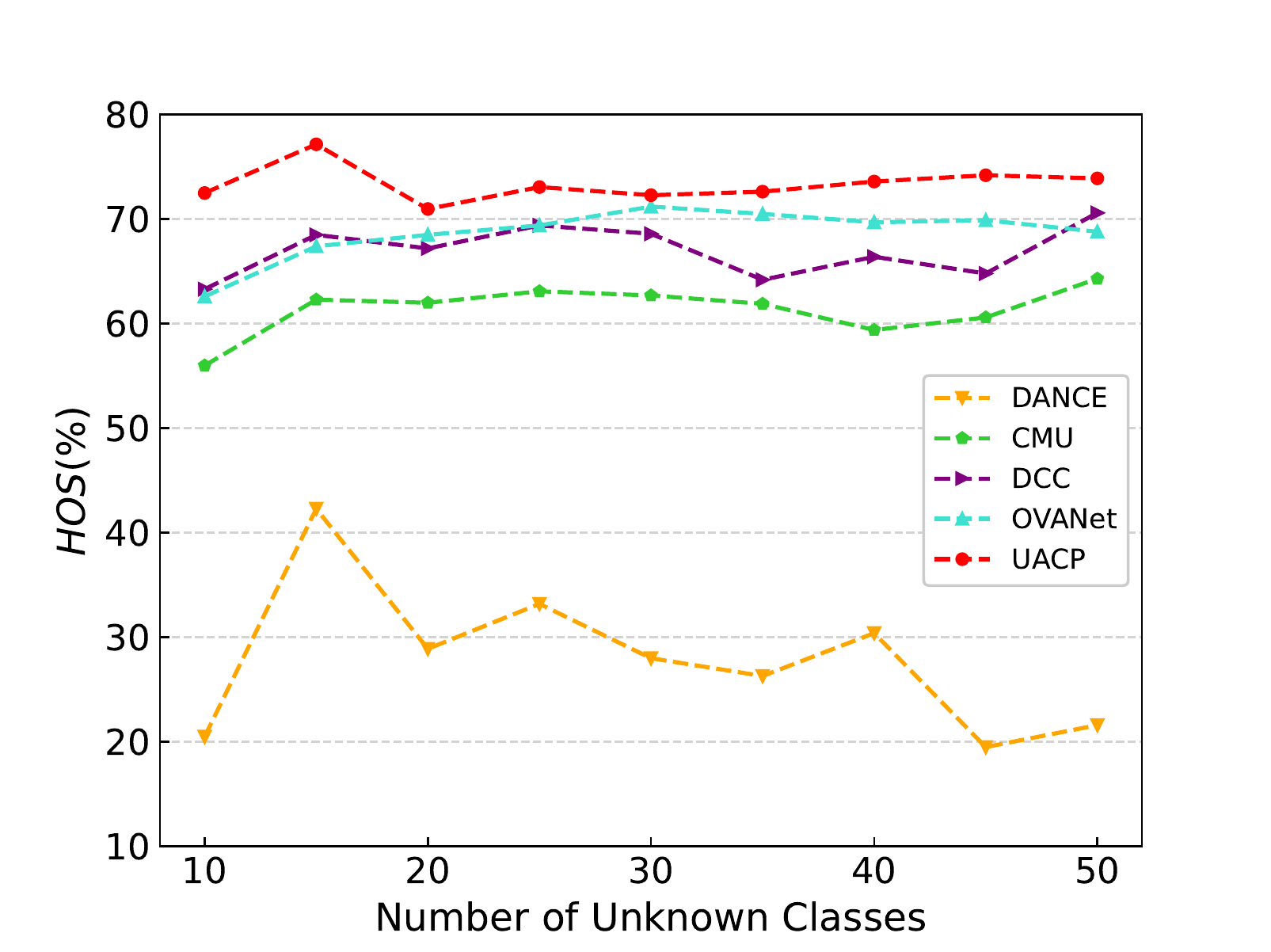}}
		\centerline{(d) Cl$\to$Pr (UniDA)}
	\end{minipage}
	\caption{$HOS$ w.r.t. varying $\lvert{L_s-L_t}\rvert$ on tasks Ar→Re and Cl→Pr for both OSDA and UniDA scenarios}
    \label{fig3}
\end{figure}
Performances by varying the number of unknown classes on 2 tasks (Ar→Re and Cl→Pr) of Office-Home in both OSDA and UniDA settings are presented in Figure~\ref{fig3}. We compare UACP with UniDA methods, including DANCE, CMU, DCC, and OVANet. Results under OSDA setting are illustrated in Figure~\ref{fig3} (a) and (b), from which it can be found that both CMU and OVANet suffer from performance degradation with the increasing number of unknown classes, while UACP has smoother fluctuations. At the same time, from comparison performance under UniDA setting in Figure~\ref{fig3}(c) and (d), UACP obviously outperforms previous state-of-the-arts with a large margin, indicating its effectiveness and robustness to different ratios of unknown classes. In summary, UACP yields consistent improvement on all tasks, demonstrating that it can effectively handle different levels of label shifts among domains.

\subsubsection{Ablation study}\label{subsubsec432}
\begin{table}[tb]
    \centering
    \begin{center}
    \begin{minipage}{\textwidth}
    \caption{Ablation on losses}\label{tab7}
    \begin{tabular*}{\textwidth}{@{\extracolsep{\fill}}lcccccc@{\extracolsep{\fill}}}
         \toprule
           & \multicolumn{3}{@{}c@{}}{Pr$\to$Re(OSDA)} &  \multicolumn{3}{@{}c@{}}{Pr$\to$Re(UniDA)}\\
         \cmidrule{2-4} \cmidrule{5-7}
           & $HOS$    & $Acc_{kn}$  & $Acc_{unk}$     & $HOS$    & $Acc_{kn}$    & $Acc_{unk}$ \\
         \midrule
           w/o $\mathcal{L}_{ESL}+\mathcal{L}_{SFC}+\mathcal{L}_{TOVA}$ & 63.5 & 70.5 & 57.9 & 71.3 & 79.4 & 64.8\\
           w/o $\mathcal{L}_{ESL}$ & 64.9 & 76.4 & 56.4 & 72.8 & 91.7 & 60.4\\
           w/o $\mathcal{L}_{SFC}$ & 72.6 & 72.5 & 72.7 & 81.4 & 90.7 & 73.9\\
           w/o $\mathcal{L}_{TOVA}$ & 70.5 & 75.4 & 66.2 & 79.1 & 92.6 & 69.1\\
           ALL & $\bf{74.8}$ & $\bf{76.7}$ & $\bf{72.9}$ & $\bf{83.3}$ & $\bf{93.3}$ & $\bf{75.2}$\\
         \botrule
    \end{tabular*}
    \end{minipage}
    \end{center}
\end{table}
In this sub-section, we verify the effectiveness of individual components of UACP. Specifically, ablation studies over the difficult task Pr→Re of Office-Home for both OSDA and UniDA settings are presented in Table~\ref{tab7}. Four variants of UACP are studied: (i) ``w/o $\mathcal{L}_{ESL}+\mathcal{L}_{SFC}+\mathcal{L}_{TOVA}$" is the variant that only trained with source supervision. (ii) ``w/o $\mathcal{L}_{ESL}$" discards classifier paradox to identify unknown samples in Eq.~(\ref{eq5}). (iii) ``w/o $\mathcal{L}_{SFC}$" discards self-supervised feature clustering on target samples in Eq.~(\ref{eq7}). (iv) ``w/o $\mathcal{L}_{TOVA}$" discards entropy minimization on OVA predictors in Eq.~(\ref{eq8}).

\begin{figure}[tb]
    \centering
	\begin{minipage}{0.4\linewidth}
		\centerline{\includegraphics[width=\textwidth]{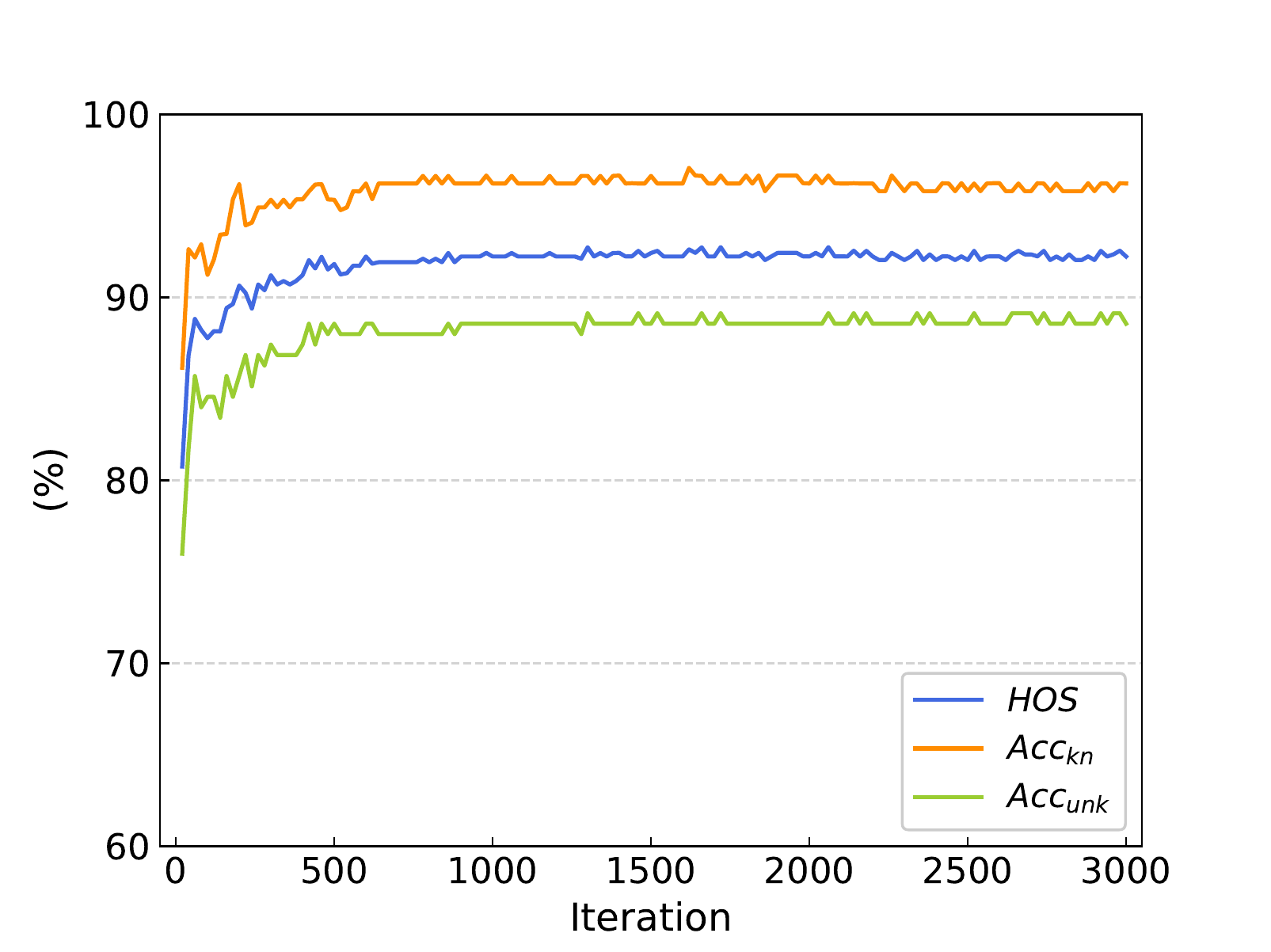}}
		\centerline{(a) A$\to$D (UACP)}
	\end{minipage}
	\begin{minipage}{0.4\linewidth}
		\centerline{\includegraphics[width=\textwidth]{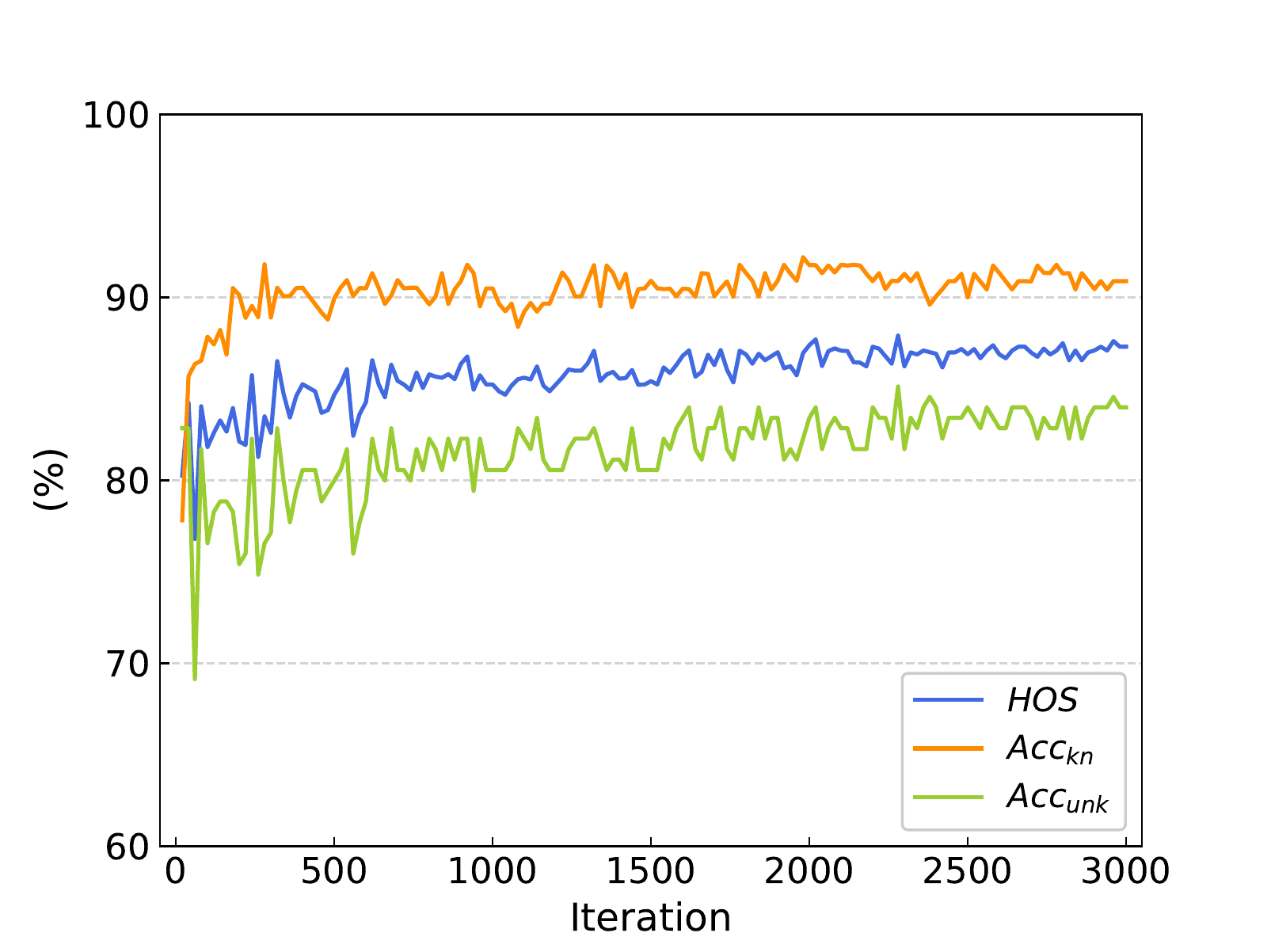}}
		\centerline{(b) A$\to$D (OVANet)}
	\end{minipage}
	\begin{minipage}{0.4\linewidth}
		\centerline{\includegraphics[width=\textwidth]{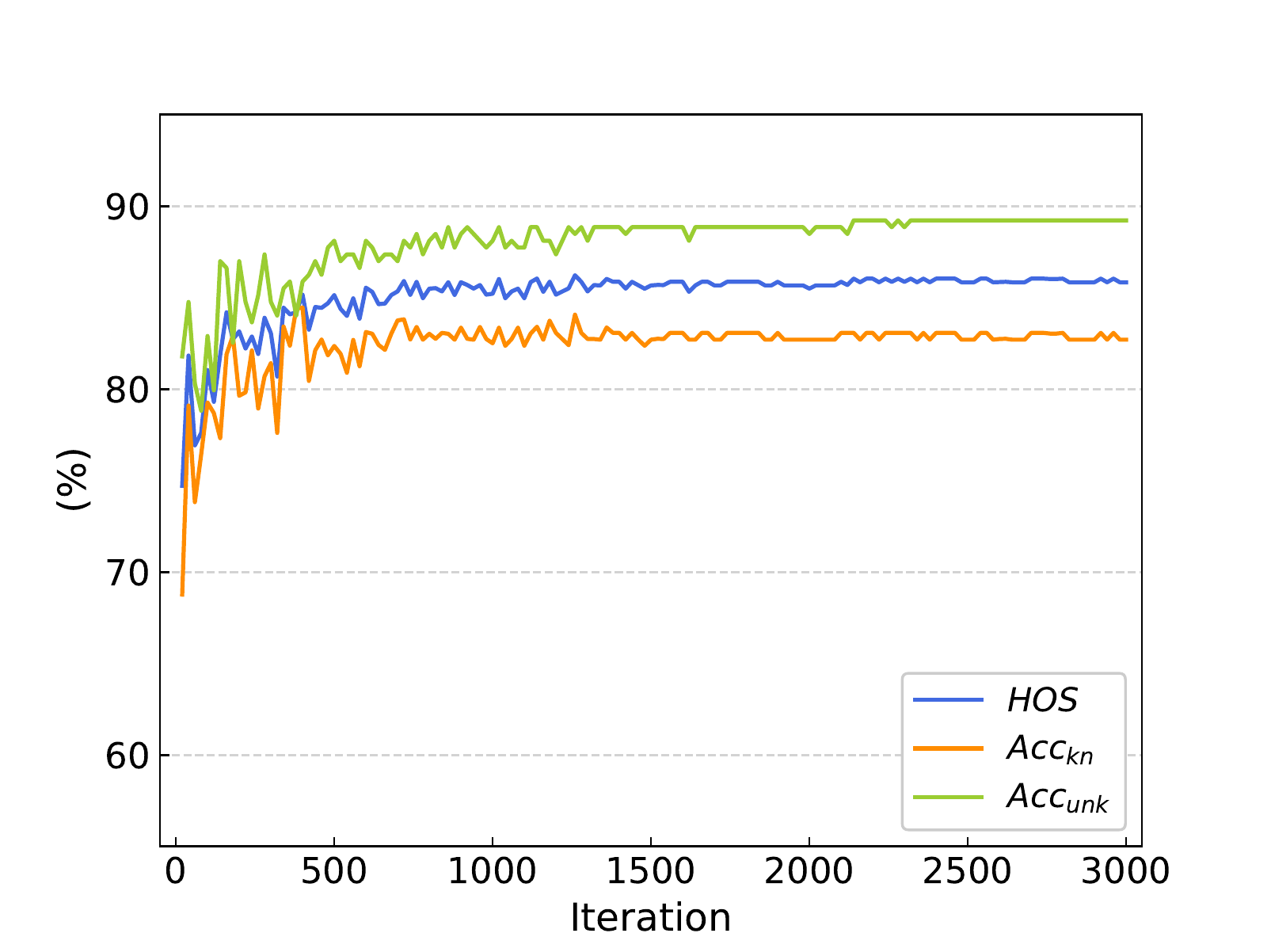}}
		\centerline{(c) A$\to$W (UACP)}
	\end{minipage}
	\begin{minipage}{0.4\linewidth}
		\centerline{\includegraphics[width=\textwidth]{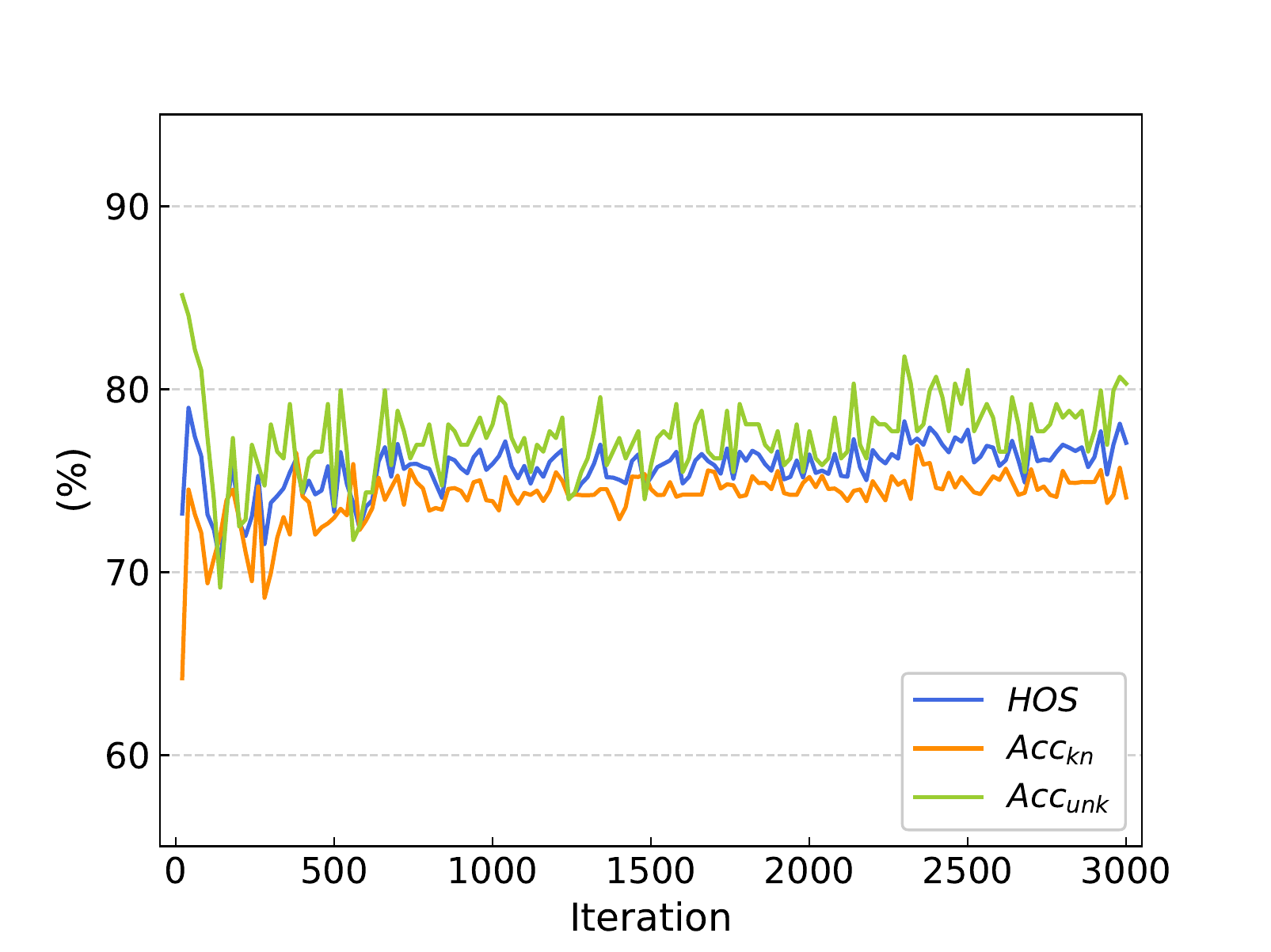}}
		\centerline{(d) A$\to$W (OVANet)}
	\end{minipage}
\caption{Comparison results of convergence speed and performance difference on UACP and OVANet}\label{fig4}
\end{figure}
\begin{figure}[tb]
    \centering
	\begin{minipage}{0.32\linewidth}
		\centerline{\includegraphics[width=\textwidth]{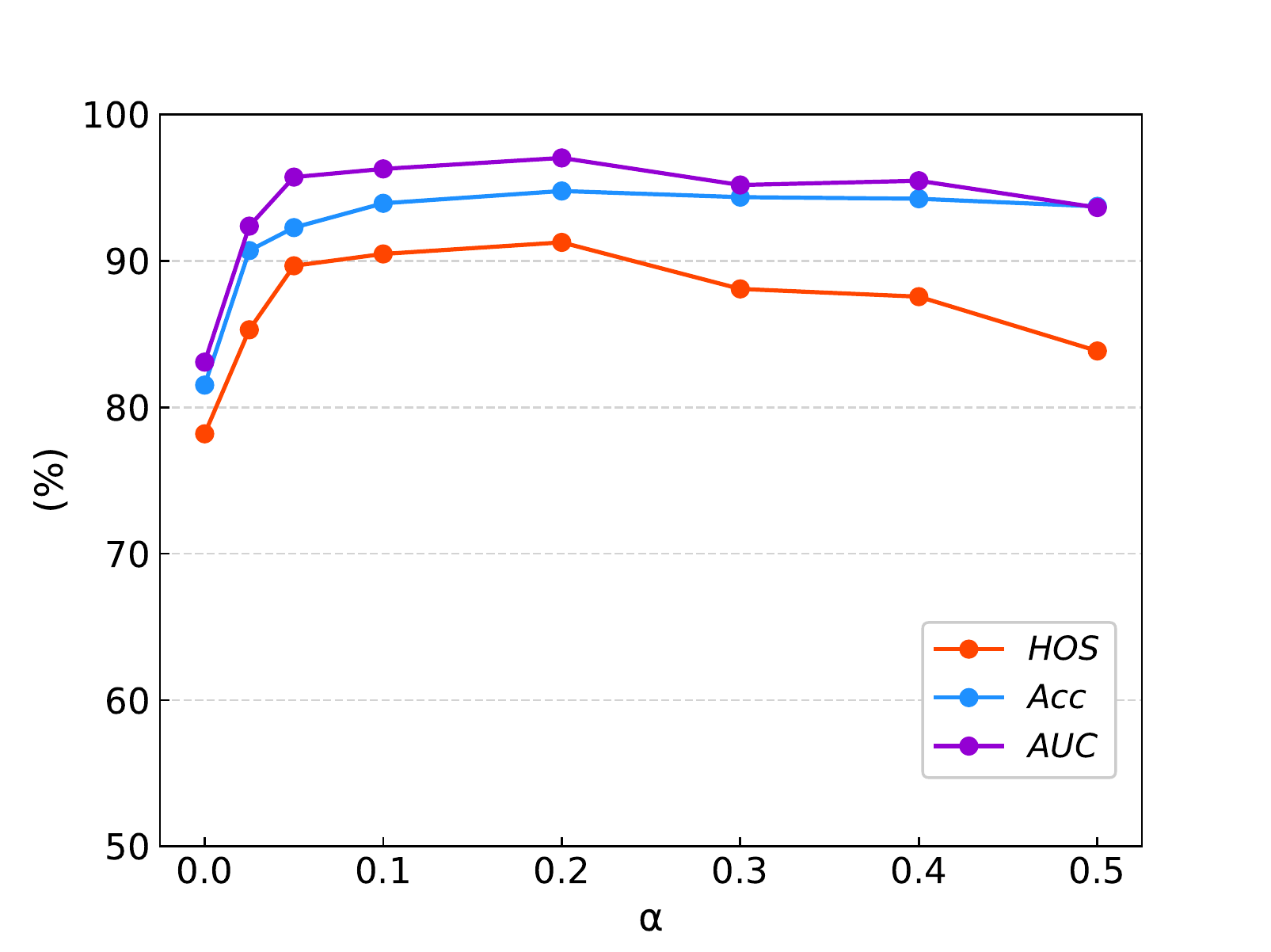}}
	\end{minipage}
	\begin{minipage}{0.32\linewidth}
		\centerline{\includegraphics[width=\textwidth]{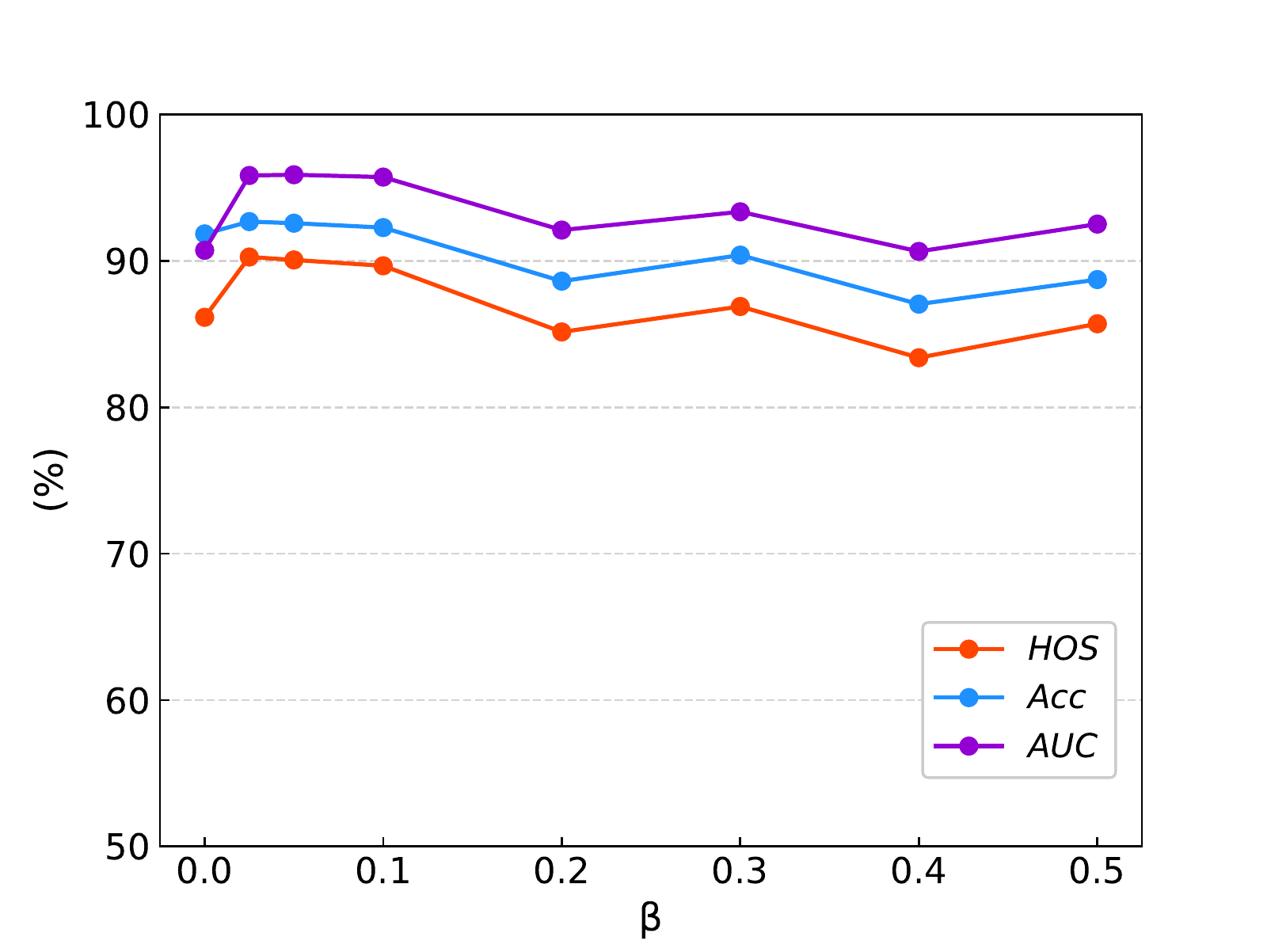}}
	\end{minipage}
	\begin{minipage}{0.32\linewidth}
		\centerline{\includegraphics[width=\textwidth]{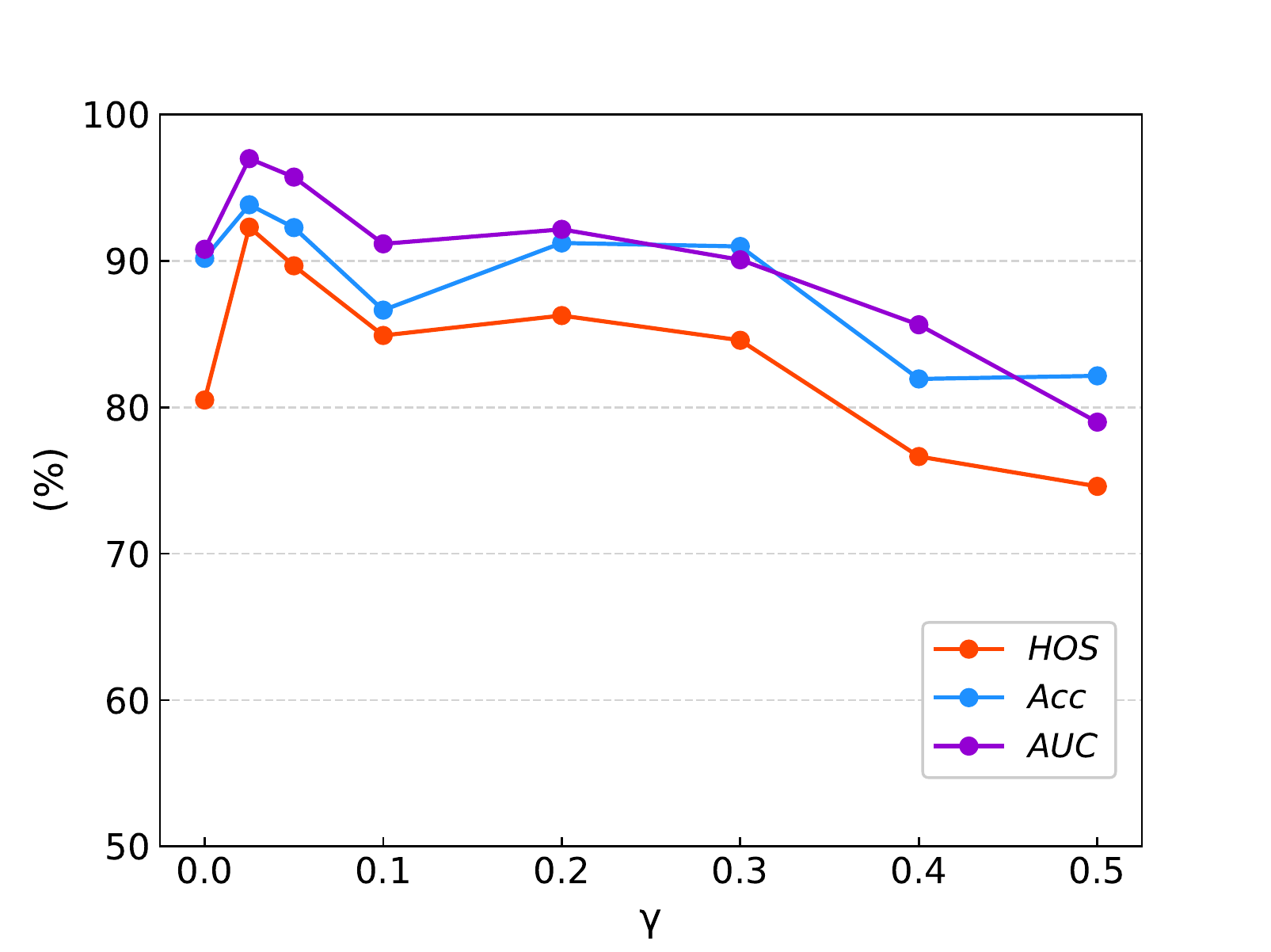}}
	\end{minipage}
	\centerline{(a) D$\to$A on Office-31}
	\begin{minipage}{0.32\linewidth}
		\centerline{\includegraphics[width=\textwidth]{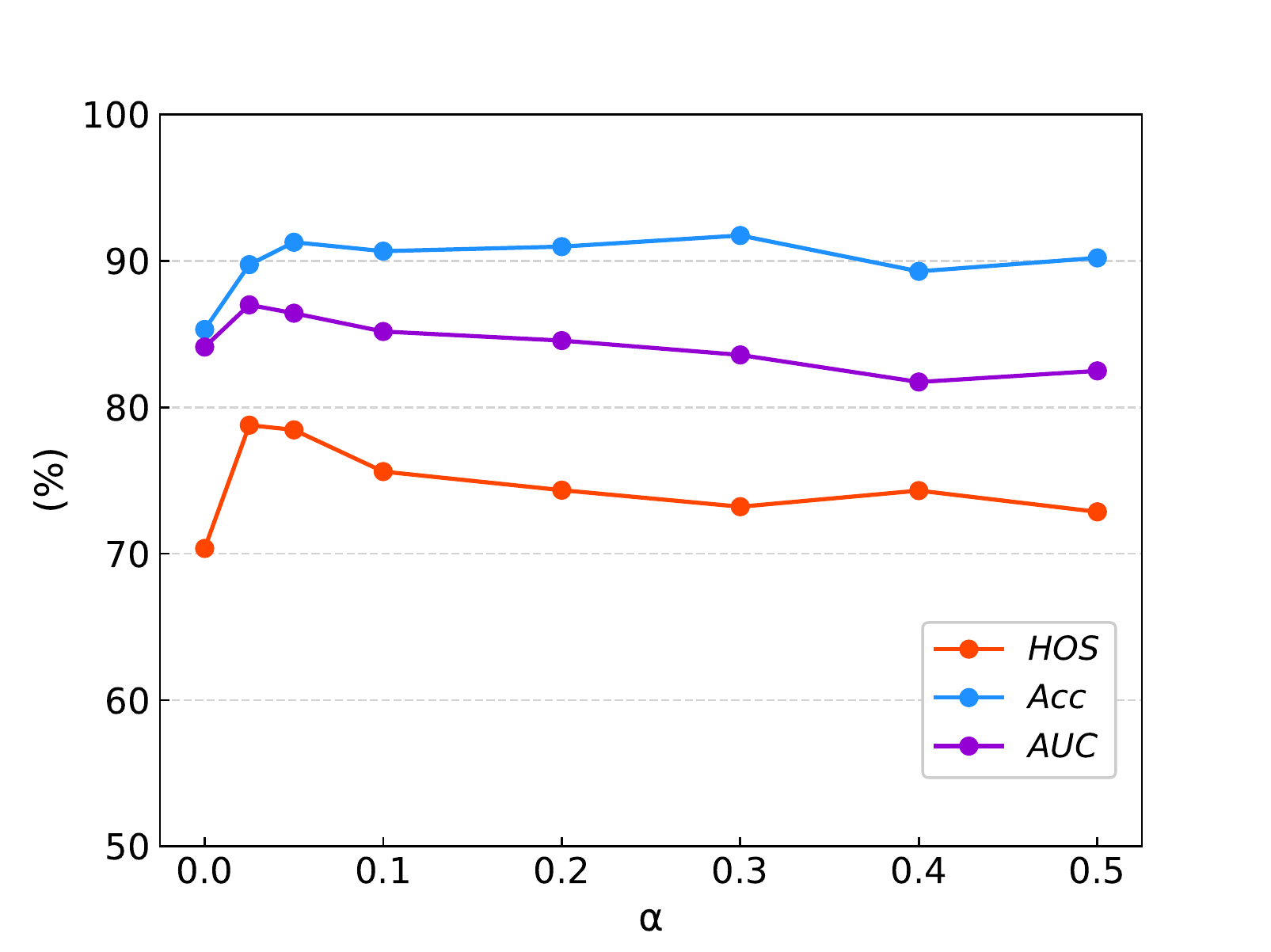}}
	\end{minipage}
	\begin{minipage}{0.32\linewidth}
		\centerline{\includegraphics[width=\textwidth]{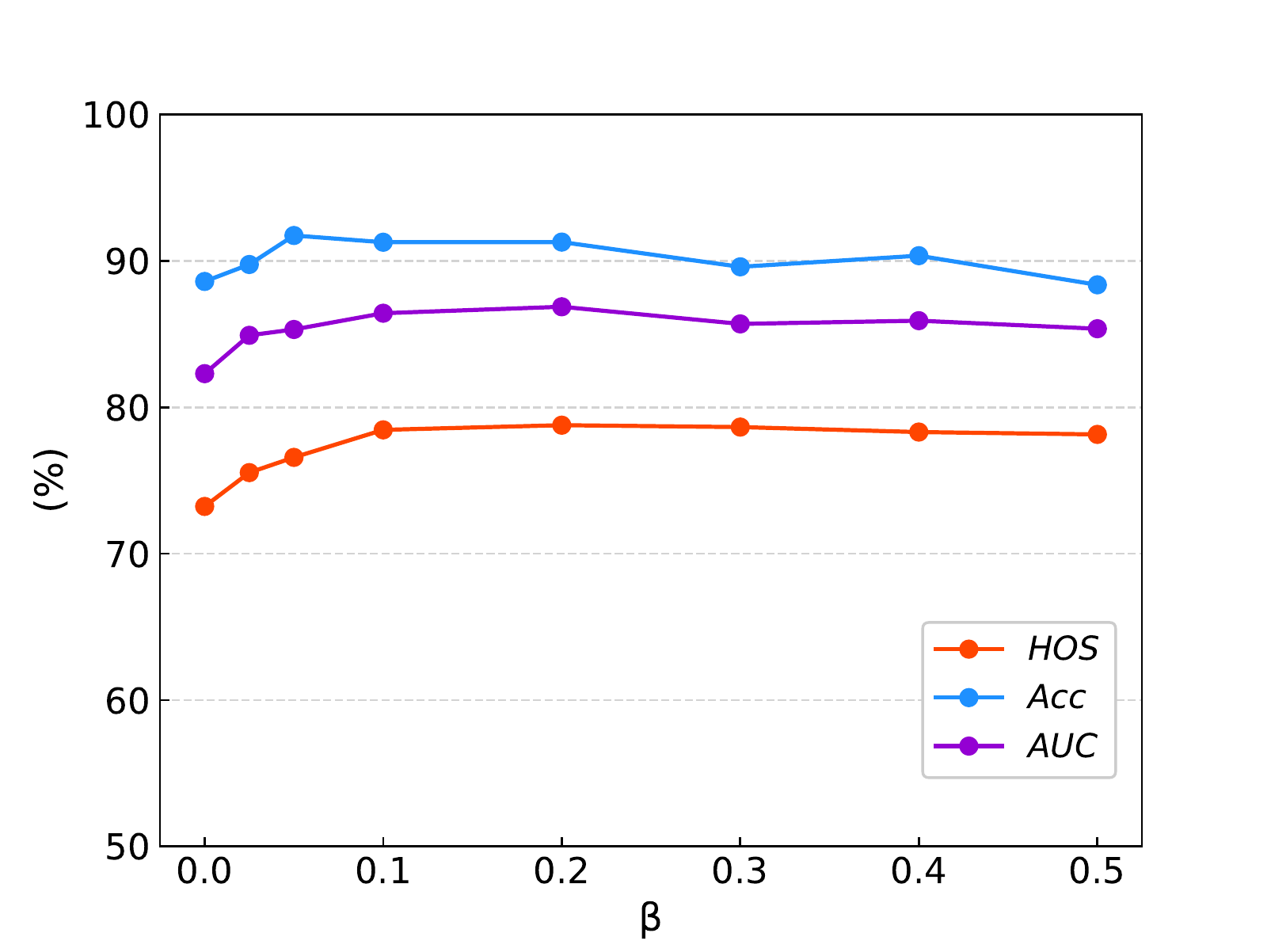}}
	\end{minipage}
	\begin{minipage}{0.32\linewidth}
		\centerline{\includegraphics[width=\textwidth]{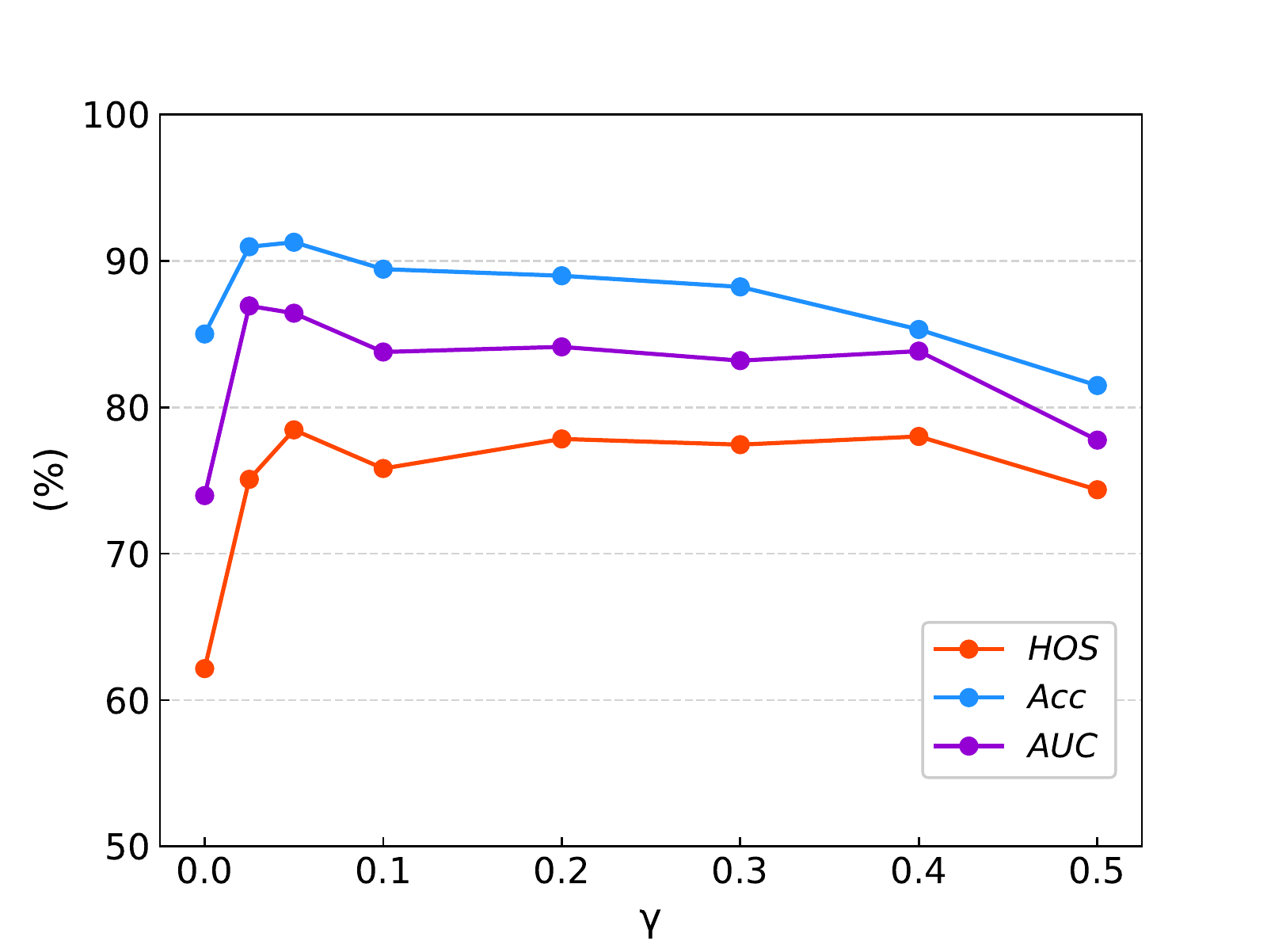}}
	\end{minipage}
	\centerline{(b) Ar$\to$Pr on Office-Home}
	\caption{ Sensitivity analysis for trade-off parameters over tasks (a) D→A of Office-31 and (b) Ar→Pr of Office-Home under UniDA scenario}
    \label{fig5}
\end{figure}
From Table~\ref{tab7}, each component of UACP contributes to the target performance. Specifically, the unknown authentication part $\mathcal{L}_{ESL}$ is essential for unknown detection, it increases the $Acc_{unk}$ significantly from 56.4\% to 72.9\% under OSDA scenario, and from 60.4\% to 75.2\% under UniDA scenario. Besides, removing the $\mathcal{L}_{SFC}$ greatly hurts the $Acc_{kn}$. When employing entropy minimization of target samples on OVA predictors, the $HOS$ is improved from 70.5\% to 74.8\% under OSDA scenario, and from 79.1\% to 83.3\% under UniDA scenario.

\subsubsection{Convergence comparison}\label{subsubsec433}

The performance of UACP in each iteration is presented in Figure~\ref{fig4}, compared to state-of-the-art OVANet. We plot $Acc_{kn}$, $Acc_{unk}$ and $HOS$ w.r.t. the number of iterations on the task A$\to$D of OSDA setting, and A$\to$W of UniDA setting, respectively. As illustrated in Figure~\ref{fig4}, UACP quickly converges within the first several hundreds of iterations and achieves better performance. Besides, our $Acc_{kn}$, $Acc_{unk}$ and $HOS$ have much less fluctuations than those of OVANet, demonstrating the stability and effectiveness of our proposal.

\subsubsection{Hyper-parameter analysis}\label{subsub434}
To illustrate the sensitivity of UACP to trade-off parameters $\alpha$, $\beta$, and $\gamma$, we perform experiments on the tasks of D$\to$A and Ar$\to$Pr under UniDA scenario. As shown in Figure~\ref{fig5}, we present the performance of $HOS$, $Acc$ and $AUC$ w.r.t. trade-off parameters $\alpha$, $\beta$, and $\gamma$ within the wide range of [0, 0.025, 0.05, 0.1, 0.2, 0.3, 0.4, 0.5]. Although the performance fluctuates as the values of parameters change to an extent, it remains relatively flat and stable within a certain range.

\section{Conclusion}
In this work, we propose a novel UniDA approach  UACP to adaptively identify unknowns by classifier paradox. In UACP, a composite classifier is proposed to tackle both domain and category shifts. The composite classifier distinguishes different source categories using MC predictor, and captures the concept of “unknown” by verification from OVA predictors. Moreover, self-supervised knowledge is utilized to pursue well-clustered target features and low-density separation for target data, so as to conduct implicit domain alignment by domain-invariant classifier. Finally, extensive experiments on four benchmarks validate UACP in diverse UDA scenarios.

\bibliography{sn-bibliography}% common bib file
% \input output

% \input sn-sample-bib.tex%

\end{document}